\documentclass[12pt,a4paper]{article}

\usepackage{graphicx} % more modern
\usepackage{subfigure} 
\usepackage{amssymb}
\usepackage{multirow}

\usepackage[numbers]{natbib}

\usepackage{algorithm}
\usepackage{algorithmic}
\usepackage{bbm}
\usepackage{amsmath}
\usepackage{color}

\usepackage[colorlinks = TRUE, citecolor = blue, urlcolor= blue]{hyperref}

\usepackage{authblk}

\title{Bayesian Consensus Clustering}
\author[1,2]{Eric F. Lock}
\author[2]{David B. Dunson}
\affil[1]{Center for Human Genetics, Duke University Medical Center, Durham, NC 27710, U.S.A.}
\affil[2]{Department of Statistical Science, Duke University, Durham, NC 27708, U.S.A.}
\date{}

\begin{document} 

\maketitle

\begin{abstract} 
The task of clustering a set of objects based on multiple sources of data arises in several modern applications.  We propose an integrative statistical model that permits a separate clustering of the objects for each data source. These separate clusterings adhere loosely to an overall consensus clustering, and hence they are not independent.  We describe a computationally scalable Bayesian framework for simultaneous estimation of both the consensus clustering and the source-specific clusterings.  We demonstrate that this flexible approach is more robust than joint clustering of all data sources, and is more powerful than clustering each data source separately.  This work is motivated by the integrated analysis of heterogeneous biomedical data, and we present an application to subtype identification of breast cancer tumor samples using publicly available data from The Cancer Genome Atlas.
Software is available at \url{http://people.duke.edu/~el113/software.html}. 
\end{abstract} 

\section{Motivation}
\label{motivation}
Several fields of research now analyze \emph{multi-source} data (also called \emph{multi-modal} data), in which multiple heterogeneous datasets describe a common set of objects.  Each dataset represents a distinct mode of measurement or domain.  Table~\ref{tab1} gives examples of multi-source data from very diverse research areas.

\begin{table*}[ht]
\begin{center}
\begin{tabular}{|l|l|p{5.8 cm}|}
\hline
Field & Object & Data sources \\
\hline
Computational biology & Tissue samples & Gene expression, microRNA, genotype, protein abundance/activity \\
\hline
Chemometrics &  Chemicals & Mass spectra, NMR spectra, atomic composition \\
\hline
Atmospheric Sciences & Locations & Temperature, humidity, particle concentrations over time \\
\hline
 Text learning & Documents & Word frequencies, authors, cited documents \\
\hline
\end{tabular}
\caption{Examples of multi-source data.  }
\label{tab1}
\end{center}
\end{table*}

While the methodology described in this article is broadly applicable, our primary motivation is the integrated analysis of heterogeneous biomedical data.  The diversity of platforms and technologies that are used to collect genomic data, in particular, is expanding rapidly.  Often multiple types of genomic data, measuring various biological components, are collected for a commen set of samples.  For example, The Cancer Genome Atlas (TCGA)  is a large-scale collaborative effort to collect and catalog data from several genomic technologies.  The integrative analysis of data from these disparate sources provides a more comprehensive understanding of cancer genetics and molecular biology.  In Section~\ref{app} we present an analysis of RNA expression, DNA methylation, microRNA expression, and proteomic data from TCGA for a common set of breast cancer tumor samples. 

Separate analyses of each data source may lack power and will not capture inter-source associations.  At the other extreme, a joint analysis that ignores the heterogeneity of the data may not capture important features that are specific to each data source.  Exploratory methods that simultaneously model shared features and features that are specific to each data source have recently been developed as flexible alternatives \cite{Lofstedt,Lock,Zhou2012,Ray}.  The demand for such integrative methods motivates a very dynamic area of statistics and machine learning.  

This article concerns integrative clustering.  Clustering is a very widely used exploratory tool to identify similar groups of objects (for example, clinically relevant disease subtypes).  Hundreds of general algorithms to perform clustering have been proposed in the literature.  However, our work is motivated by the need for an integrative clustering method that is computationally scalable and robust to the unique features of each data source.    

\section{Related Work}
\label{work}
\subsection{Integrative clustering}
Most applications of clustering multi-source data follow one of two general approaches:
\begin{enumerate}
\item Clustering of each data source separately,  potentially followed by a post hoc integration of these separate clusterings.  
\item Combining all data sources to determine a single ``joint" clustering.  
\end{enumerate}  
Under approach (1) the level of agreement between the separate clusterings may be measured by the \emph{adjusted rand index} \cite{Hubert} or a similar statistic.  Furthermore, \emph{consensus clustering} can be used to determine an overall partition of the objects that agrees the most with the data-specific clusterings.  Several objective functions and algorithms to perform consensus clustering have been proposed (for a survey see \citet{Nguyen}).  These methods are most commonly used to establish consensus among multiple clustering algorithms, or multiple realizations of the same clustering algorithm, on a single dataset.  Consensus clustering has also been used to integrate multi-source data \cite{Filkov,Perou}.  Such an approach is attractive in that it models source-specific features yet still determines an overall clustering, which is often of practical interest.  However, the two stage process of performing entirely separate clusterings followed by post hoc integration limits the power to identify and exploit shared structure  (see Section~\ref{ClusterComp} for an illustration of this phenomenon).  

Approach (2) effectively exploits shared structure, at the expense of failing to recognize features that are specific to each data source.  Within a model-based statistical framework, one can find the clustering that maximizes a joint likelihood.  Assuming that each source is conditionally independent given the clustering, the joint likelihood is the product of the likelihood functions for each data source.   This approach is used by \citet{Kormaksson} in the context of integrating gene expression and DNA methylation data.  The \emph{iCluster} method \cite{Shen,Mo2013} performs clustering by first fitting a Gaussian latent factor model to the joint likelihood; clusters are then determined by K-means clustering of the factor scores.  \citet{Rey} propose a dependency-seeking model in which the goal is to find a clustering that accounts for associations across the data sources.  

Perhaps most similar to our approach in spirit and motivation is the \emph{Multiple Dataset Integration} (MDI) method \cite{Kirk2012}, which uses a statistical framework to cluster each data source separately while simultaneously modeling dependence between the clusterings.  By explicitly modeling dependence, MDI permits borrowing strength across data sources.  The fundamental difference between MDI and the approach we describe in the following concerns how dependence is modeled.  Specifically, MDI models the dependence between each pair of data sources rather than adherence to an overall clustering.  We elaborate on this distinction in Sections~\ref{MargForms} and~\ref{ClusterComp}.       

\subsection{Finite Dirichlet mixture models}  
Here we briefly describe the finite Dirichlet mixture model for clustering a single dataset, with the purpose of laying the groundwork for the integrative model given in Section~\ref{model}.  Given data $X_n$ for $N$ objects ($n=1,...,N$), the goal is to partition these objects into at most $K$ clusters.  Typically $X_n$ is a multi-dimensional vector, but we present the model in sufficient generality to allow for more complex data structures.  Let $f(X_n|\theta)$ define a probability model for $X_n$ given parameter(s) $\theta$.  For example, $f$ may be a Gaussian density defined by the mean and variance $\theta = (\mu, \sigma^2)$.  Each $X_n$ is drawn independently from a mixture distribution with $K$ components, specified by the parameters $\theta_1,\hdots,\theta_K$.  Let $C_n \in \{1,\hdots,K\}$ represent the component corresponding to $X_n$, and $\pi_k$ be the probability that an arbitrary object belongs to cluster $k$: 
\[ \pi_k = P(C_n= k). \]
 Then, the generative model is 
\[X_n \sim f(\cdot|\theta_{k}) \text{ with probability } \pi_k.\]
We further assume that $\Pi = (\pi_1,\hdots,\pi_K)$ has a Dirichlet distribution parameterized by a $K$-dimensional vector $\beta$ of positive reals.  This allows some $\pi_k$ to be small, and therefore $N$ objects may not represent all $K$ clusters.    Letting $K \rightarrow \infty$ gives a \emph{Dirichlet process}.  

Under a Bayesian framework one can put a prior distribution on $\Pi$ and the parameter set $\Theta = (\theta_1,\hdots,\theta_K)$. Standard  computational methods such as Gibbs sampling can then be used to approximate the posterior distribution for $\Pi$, $\Theta$, and $\mathbb{C} = (C_1,\hdots,C_N)$ (for an overview see \citet{Neal}).         
          
\section{Integrative model}
\label{model}
We extend the Dirichlet mixture model to accommodate data from $M$ sources $\mathbb{X}_1,\hdots,\mathbb{X}_M$.  Each data source is available for a common set of $N$ objects,  where $X_{mn}$ represents data $m$ for object $n$.  Each data source requires a probability model $f_m(X_n|\theta_m)$ parametrized by $\theta_m$.  Under the general framework presented here each $\mathbb{X}_m$ may have disparate structure.  For example $X_{1n}$ may give an image where $f_1$ defines the spectral density for a Gaussian random field, while $X_{2n}$ may give a categorical vector where $f_2$ defines a multivariate probability mass function. 

We assume there is a separate clustering of the objects for each data source, but that these adhere loosely to an overall clustering. Formally, each $X_{mn}$ $n=1,\hdots,N$ is drawn independently from a $K$-component mixture distribution specified by the parameters $\theta_{m1},\hdots,\theta_{mK}$.  Let $L_{mn} \in \{1,\hdots,K\}$ represent the component corresponding to $X_{mn}$.  Furthermore, let $C_{n} \in \{1,\hdots,K\}$ represent the overall mixture component for object $n$.  The source-specific clusterings $\mathbb{L}_m = (L_{m1},\hdots,L_{mN})$ are dependent on the overall clustering $\mathbb{C} = (C_{1},\hdots,C_{N})$:
\[P(L_{mn} = k|C_n) = \nu(k,C_n, \alpha_m) \]
where $\alpha_m$ adjusts the dependence function $\nu$.  The data $\mathbb{X}_m$ are independent of $\mathbb{C}$ conditional on the source-specific clustering $\mathbb{L}_m$.  Hence, $\mathbb{C}$ serves only to unify $\mathbb{L}_1,\hdots,\mathbb{L}_M$.  The conditional model is 
\[P(L_{mn} = k |X_{mn}, C_n, \theta_{mk}) \propto \nu(k,C_n, \alpha_m) f_m(X_{mn} | \theta_{mk}).\]
 
Throughout this article, we assume $\nu$ has the simple form
\begin{eqnarray} \nu(L_{mn},C_n, \alpha_m) =  \left\{ \begin{array}{c} \alpha_m \text{ if } C_n = L_{mn} \\ \frac{1-\alpha_m}{K-1} \text{ otherwise} \end{array} \right. \label{nuDef} \end{eqnarray}
where $\alpha_m \in [\frac{1}{K},1]$ controls the adherence of data source $m$ to the overall clustering.  More simply $\alpha_m$ is the probability that $L_{mn} = C_n$. So, if $\alpha_m = 1$ then $\mathbb{L}_m = \mathbb{C}$.  The $\alpha_m$ are estimated from the data together with $\mathbb{C}$ and $\mathbb{L}_1,\hdots,\mathbb{L}_m$.  In practice we estimate each $\alpha_m$ separately, or assume that $\alpha_1=\hdots=\alpha_M$ and hence each data source adheres equally to the overall clustering.  The latter is favored when $m=2$ for identifiability reasons.  More complex models that permit dependence of the $\alpha_m$'s are also potentially useful.    

Let $\pi_k$ be the probability that an object belongs to the overall cluster $k$:
\[\pi_k = P(C_n = k).\]
We assume $\Pi = (\pi_1,\hdots,\pi_K)$ follows a Dirichlet($\beta$) distribution.  The probability that an object belongs to a given source-specific cluster follows directly:
\begin{eqnarray} P(L_{mn}=k|\Pi) = \pi_k \alpha_m+(1-\pi_k) \frac{1-\alpha_m}{K-1}. \label{Ldist} \end{eqnarray}
Moreover, a simple application of Bayes rule gives the conditional distribution of $\mathbb{C}$:
\[P(C_n=k|\mathbb{L},\Pi,\mathbb{\alpha}) \propto \pi_k \prod_{m=1}^M \nu(L_{mn},k,\alpha_m), \]
where $\nu$ is defined as in (\ref{nuDef}).    

Although the number of possible clusters $K$ is the same for $\mathbb{L}_1,\hdots,\mathbb{L}_M$ and $\mathbb{C}$, the number of clusters that are actually represented may vary.  Generally the source-specific clusterings $\mathbb{L}_m$ will represent more clusters than $\mathbb{C}$, rather than vice-versa.  This follows from Equation (\ref{Ldist}) and is illustrated in Appendix\~ref{Sec2}.  Intuitively if object $n$ is not allocated to any overall cluster in data source $m$ (i.e., $L_{mn} \notin \mathbb{C}$) then $X_{mn}$ does not conform well to any overall pattern in the data.

\begin{table}[ht]
\begin{center}
\begin{tabular}{|l|l|p{2.5 cm}|}
\multicolumn{2}{c}{\textbf{Notation}}\\
\hline
$N$ & Number of objects \\
$M$ & Number of data sources \\
$K$ & Number of clusters \\
$\mathbb{X}_m$ & Data source $m$\\
$X_{mn}$ & Data for object $n$, source $m$  \\
$f_m$ & Probability model for source $m$ \\
$\theta_{mk}$ & Parameters for $f_m$, cluster $k$ \\
$p_{m}$ & Prior distribution for $\theta_{mk}$\\
$C_n$ & Overall cluster for object $n$ \\
$\pi_k$ & Probability that $C_n = k$ \\
$L_{mn}$ & Cluster specific to $X_{mn}$ \\
$\nu$ & Dependence function for $C_n$ and $L_{mn}$ \\ 
$\alpha_m$ & Probability that $L_{mn} = C_n$ \\
\hline
\end{tabular}
\caption{Notation.}
\label{tab2}
\end{center}
\end{table}

\subsection{Marginal forms}
\label{MargForms}
Integrating over the overall clustering $C$ gives the joint marginal distribution of $\mathbb{L}_{1} ,\hdots,\mathbb{L}_M$:    
\begin{eqnarray}P(\{L_{mn}=k_m\}_{m=1}^M|\Pi,\mathbb{\alpha}) \propto \sum_{k=1}^K \pi_k \prod_{m=1}^M \nu(k_m,k,\alpha_m). \label{GenMarg} \end{eqnarray}
Under the assumption that $\alpha_1=\hdots=\alpha_M$ the model simplifies: 
\begin{eqnarray}P(\{L_{mn}=k_m\}_{m=1}^M|\Pi,\mathbb{\alpha}) \propto \sum_{k=1}^K \pi_k U^{t_k} \label{Uform}\end{eqnarray}
where $t_k$ is the number of clusters equal to $k$ and $U = \frac{(K-1) \alpha_1}{1-\alpha_1} \geq 1$. 
 This marginal form facilitates comparison with the MDI method for dependent clustering.  In the MDI model $\phi_{ij}>0$ control the strength of association between the clusterings $\mathbb{L}_i$ and $\mathbb{L}_j$:
\begin{eqnarray}P(\{L_{mn}=k_m\}_{m=1}^M|\tilde{\Pi},\Phi) \propto  \prod_{m=1}^{M}  \tilde{\pi}_{mk_m} \hspace{-13 pt} \prod_{\{i<j|k_i=k_j\}} \hspace{-10 pt} (1 + \phi_{ij}) \label{MDIform} \end{eqnarray}
where $\tilde{\pi}_{mk} = P(L_{mn} = k)$.  For $M =2$ and $\tilde{\pi}_{1 \cdot} = \tilde{\pi}_{2 \cdot}$ it is straightforward to show that (\ref{Uform}) and (\ref{MDIform}) are functionally equivalent under a parameter substitution (see Appendix~\ref{Sec3}).  There is no such equivalence for $M>2$, regardless of restrictions on $\tilde{\Pi}$ and $\Phi$.  This is not surprising, as MDI gives a general model of pairwise dependence between clusterings rather than a model of adherence to an overall clustering. 

\section{Computation} \label{Comp}
Here we present a general Bayesian framework for estimation of the integrative clustering model.   We employ a Gibbs sampling procedure to approximate the posterior distribution for the parameters introduced in Section~\ref{model}.  The algorithm is general in that we do not assume any specific form for the $f_m$ and the parameters $\theta_{mk}$.  We use conjugate prior distributions for $\alpha_m$, $\Pi$, and (if possible) $\theta_{mk}$.   
\begin{itemize}
\item $\alpha_m \sim \text{TBeta}(a_m,b_m,\frac{1}{K})$,  the $\text{Beta}(a_m,b_m)$ distribution truncated below by $\frac{1}{K}$.  By default we choose $a_m=b_m=1$, so that the prior for $\alpha_m$ is uniformly distributed between $\frac{1}{K}$ and $1$.     
\item $\Pi \sim \text{Dirichlet}(\beta_0)$.  By default we choose $\beta_0 = (1,1,...,1)$, so that the prior for $\Pi$ is uniformly distributed on the standard $(M-1)$-simplex.      
\item The $\theta_{mk}$ have prior distribution $p_m$.  In practice, one should choose $p_m$ so that sampling from the conditional posterior $p_m(\theta_{mk}| \mathbb{X}_m, \mathbb{L}_m)$ is feasible.  
\end{itemize}
Markov chain Monte Carlo (MCMC) proceeds by iteratively sampling from the following conditional posterior distributions: 
\begin{itemize}
\item $\Theta_m |  \mathbb{X}_m, \mathbb{L}_m   \sim  p_m(\theta_{mk}| \mathbb{X}_m, \mathbb{L}_m)$ for $k=1,\hdots,K$.
\item $\mathbb{L}_{m} | \mathbb{X}_m,\Theta_m,\alpha_m,\mathbb{C} \sim P(k |X_{mn}, C_n, \theta_{mk},\alpha_m)$ for $n=1,\hdots,N$, where 
\[P(k|X_{mn}, C_n, \Theta_{m}) \propto \nu(k,C_n, \alpha_m) f_m(X_{mn} | \theta_{mk}).\]
\item $\alpha_m | \mathbb{C},\mathbb{L}_m \sim \text{TBeta}(a_m+\tau_m, b_m+N-\tau_m,\frac{1}{K}),$ where $\tau_m$ is the number of samples $n$ satisfying $L_{mn} = C_n$. 
\item $\mathbb{C}|\mathbb{L}_{m},\Pi,\alpha \sim  P(k |\Pi,\{L_{mn},\alpha_m\}_{m=1}^M)$ for $n=1,\hdots,N$, where
 \[P(k |\Pi,\{L_{mn},\alpha_m\}_{m=1}^M) \propto \pi_k \prod_{m=1}^M \nu(k,L_{mn},\alpha_m)\]
\item $\Pi | \mathbb{C} \sim \text{Dirichlet}(\beta_0+\rho)$, where $\rho_k$ is the number of samples allocated to cluster $k$ in  $\mathbb{C}$.
\end{itemize}
This algorithm can be suitably modified under the assumption that $\alpha_1=\hdots=\alpha_M$ (see Appendix~\ref{appendix1.2}). 

 Each sampling iteration produces a different realization of the clusterings $\mathbb{C}, \mathbb{L}_1,\cdots,\mathbb{L}_m$, and together these samples approximate the posterior distribution for the overall and source-specific clusterings.  However, a point estimate may be desired for each of  $\mathbb{C}, \mathbb{L}_1,\cdots,\mathbb{L}_m$ to facilitate interpretation of the clusters.   In this respect methods that aggregate over the MCMC iterations to produce a single clustering, such as that described in \cite{Dahl}, can be used.  

It is possible to derive a similar sampling procedure using only the marginal form for the source-specific clusterings given in Equation~(\ref{GenMarg}). However, the overall clustering $C$ is also of interest in most applications.  Furthermore, incorporating $C$ into the algorithm can actually improve computational efficiency dramatically, especially if $M$ is large.  As presented, each MCMC iteration can be completed in $O(MNK)$ operations.  If the full joint marginal distribution of $L_1,\hdots,L_M$ is used the computational burden increases exponentially with $M$ (this also presents a bottleneck for the MDI method).  

For each iteration, $C_n$ is determined randomly from a distribution that gives higher probability to clusters that are prevalent in $\{L_{1n},\hdots,L_{mn}\}$.  In this sense $\mathbb{C}$ is determined by a random consensus clustering of the source-specific clusterings.  Hence, we refer to this approach as \emph{Bayesian consensus clustering} (BCC).  BCC differs from traditional consensus clustering in three key aspects. 
\begin{enumerate}
\item Both the source-specific clusterings and the consensus clustering are modeled in a statistical way that allows for uncertainty in all parameters.  
\item The source-specific clusterings and the consensus clustering are estimated simultaneously, rather than in two stages.  This permits borrowing of information across sources for more accurate cluster assignments.  
\item The strength of association to the consensus clustering for each data source is learned from the data and accounted for in the model.
\end{enumerate}

We have developed software for the R environment for statistical computing \cite{R} to perform BCC on multivariate continuous data sources (available at \url{http://people.duke.edu/~el113/software.html}).  Specifically, we use a Normal-Gamma conjugate prior distribution to model cluster-specific means and variances.  See Appendix~\ref{Sec1} for more details.  

\subsection{Choice of $K$}
\label{chooseK}
In theory the specified maximum number of clusters $K$ can be large, for example $K=N$.  The number of clusters realized in $\mathbb{C}$ and the $\mathbb{L}_m$ may still be small.  However, we find that this is not the case for high-dimensional structured data such as that used for the genomics application in Section~\ref{app}.  The model tends to select a large number of clusters even if the Dirichlet prior concentration parameters $\beta_0$ are very small.  From an exploratory perspective we would like to identify a small number of interpretable clusters.  For this reason Dirichlet process models that allow for an infinite number of clusters are also not appropriate.     

We therefore consider an alternative heuristic measure that selects the value of $K$ that gives maximum adherence to an overall clustering.  The adjusted adherence parameters $\alpha_m^* \in [0,1]$ given by  
\[\alpha_m^* = \frac{K\alpha_m-1}{K-1}\]
are computed for each candidate value of $K$.  We then select the $K$ giving the highest mean adjusted adherence
\[\bar{\alpha}^* = \frac{1}{M}\sum_{m=1}^M \alpha_m^*  .\]
In practice, we find that this method selects a small number of clusters that reveal shared structure across the data sources.

\section{Simulation}
\label{sims}
Here we present applications of BCC to simple simulated datasets.  To the best of our knowledge there is no existing method that is directly comparable to BCC in terms of the motivating model.  Nevertheless, we do illustrate the flexibility and potential advantages of BCC over other model-based clustering methods for multi-source data.    

\subsection{Accuracy of $\hat{\alpha}$}
\label{AlphaAcc}
We find that with reasonable signal the $\alpha_m$ can generally be estimated with accuracy and without substantial bias.  To illustrate, we generate simulated datasets $\mathbb{X}_1: 1 \times 200$ and $\mathbb{X}_2:1 \times 200$ as follows:
\begin{enumerate}
\item Let $\mathbb{C}$ define two clusters, where $C_n = 1$ for $n \in \{1,\hdots,100\}$ and $C_n = 2$ for $n \in \{101,\hdots,200\}$ 
\item Draw $\alpha$ from a Uniform$(0.5,1)$ distribution.
\item For $m=1,2$ and $n=1,\hdots,200$, generate $L_{mn} \in \{1,2\}$ with probabilities $P(L_{mn} = C_n) = \alpha$ and $P(L_{mn} \neq C_n) = 1-\alpha$.
\item For $m=1,2$ draw values $X_{mn}$ from a Normal$(1.5,1)$ distribution if $L_{mn}=1$ and from a Normal$(-1.5,1)$ distribution if $L_{mn}=2$ 
\end{enumerate}

 We generate 100 realizations of the above simulation, and estimate the model via BCC for each realization.  We assume $\alpha_1=\alpha_2$ in our estimation and use a uniform prior; further computational details are given in Section 4 of the supplemental article.  Figure~\ref{fig:alphaEsts} displays $\hat{\alpha}$, the best estimate for both $\alpha_1$ and $\alpha_2$, versus the true $\alpha$ for each realization.  The point estimate displayed is the mean over MCMC draws, and we also display a 95\% credible interval based on the 2.5 of 97.5 percentiles of the MCMC draws.  The estimated $\hat{\alpha}$ are generally close to $\alpha$, and the credible interval contains the true value in 91 of 100 simulations. See Section 4 of the supplemental article for a more detailed study, including a simulation illustrating the effect of the prior distribution on $\hat{\alpha}$. 

\begin{figure}[!ht]
\vskip 0.2in
\begin{center}
\centerline{\includegraphics[width=0.75\columnwidth, trim = 0mm 5mm 0mm 7mm, clip = TRUE]{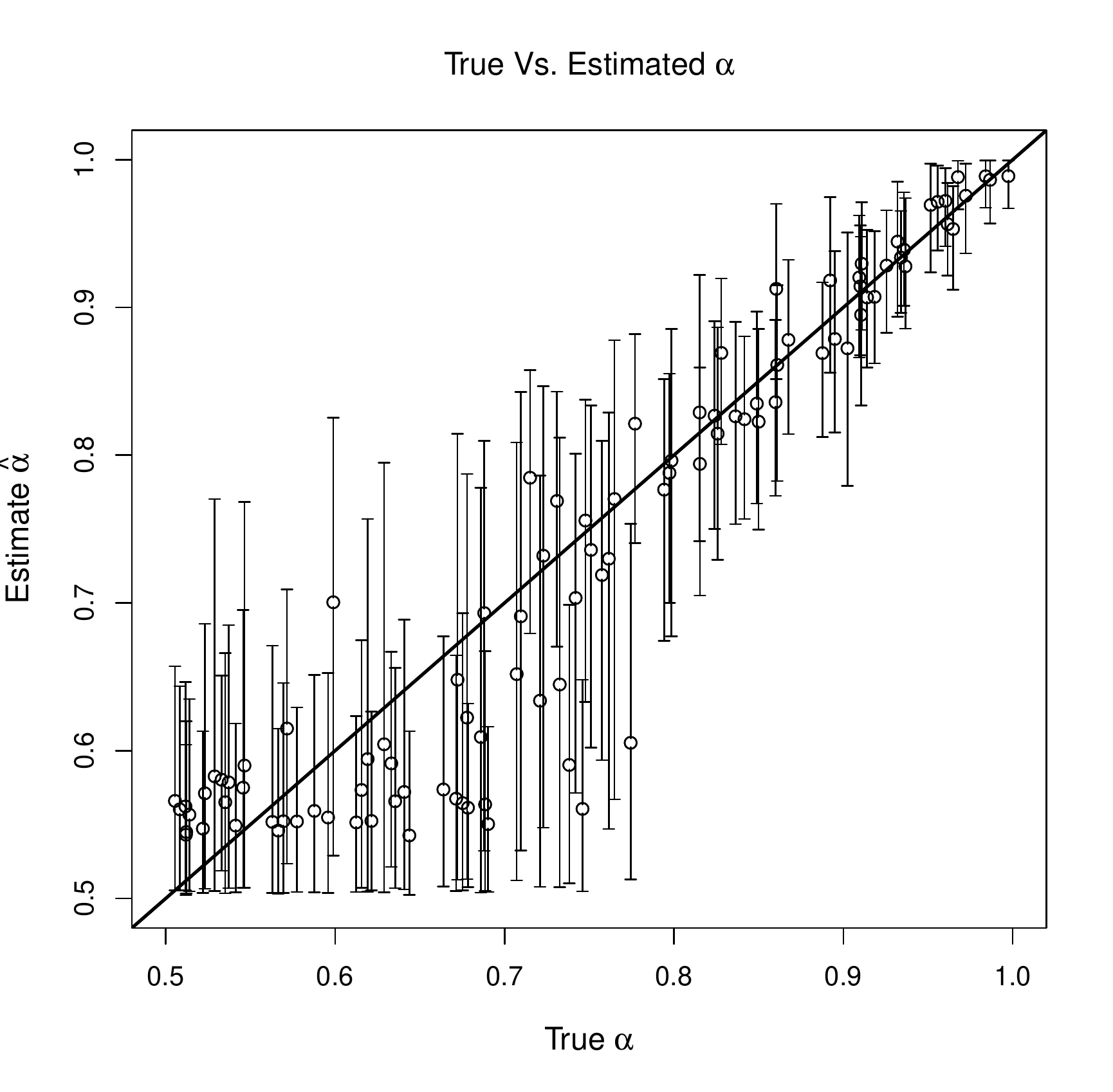}}
\caption{Estimated $\hat{\alpha}$ vs. true $\alpha$ for 100 randomly generated simulations.  For each simulation the mean value $\hat{\alpha}$ is shown with a 95\% credible interval. }
\label{fig:alphaEsts}
\end{center}
\vskip -0.2in
\end{figure}

\subsection{Clustering accuracy}
\label{ClusterComp}
To illustrate the flexibility and advantages of BCC in terms of clustering accuracy, we generate simulated data sources $\mathbb{X}_1$ and $\mathbb{X}_2$ as in Section~\ref{AlphaAcc} but with Normal(1,1) and Normal(-1,1) as our mixture distributions.  Hence, the signal distinguishing the two clusters is weak enough so that there is substantial overlap within each data source.  We generate 100 simulations and compare the results for four model-based clustering approaches:
\begin{enumerate}
\item Separate clustering, in which a finite Dirichlet mixture model is used to determine a clustering separately for $\mathbb{X}_1$ and $\mathbb{X}_2$. 
\item Joint clustering, in which a finite Dirichlet mixture model is used to determine a single clustering for the concatenated data $\left[\begin{array}{c}\mathbb{X}_1 \\ \mathbb{X}_2 \end{array} \right ]$.
\item Dependent clustering, in which we model the pairwise dependence between each data source, in the spirit of MDI.
\item Bayesian consensus clustering.
\end{enumerate}
The full implementation details for each method are given in Section 5 of the supplemental article.  

\begin{figure}[!ht]
\vskip -0.15in
\begin{center}
\subfigure
{\label{fig:M2}
\includegraphics[width=0.65\columnwidth]{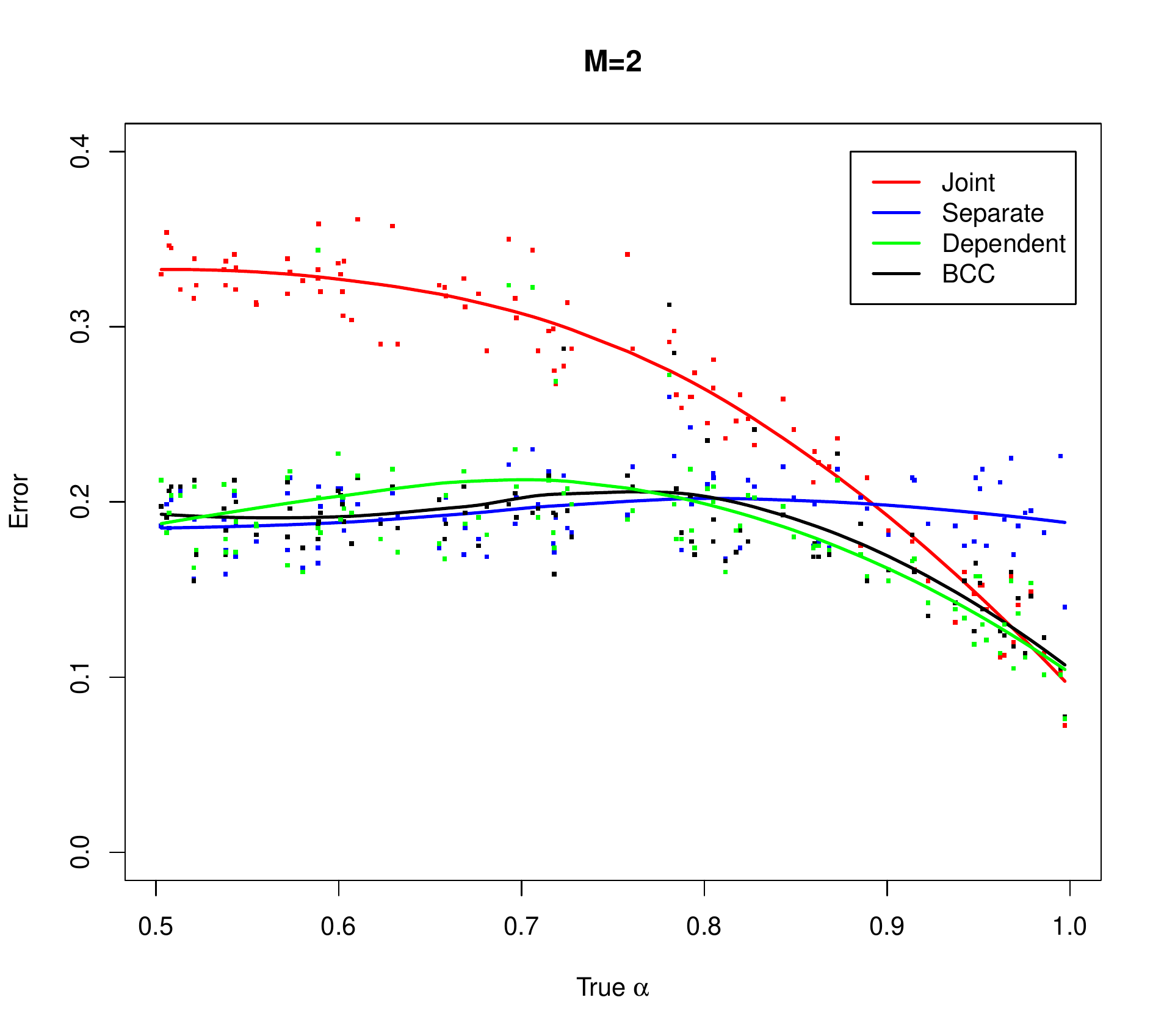}}
%\vskip -0.2in
\subfigure{
\label{fig:M3}
\includegraphics[width=0.65\columnwidth]{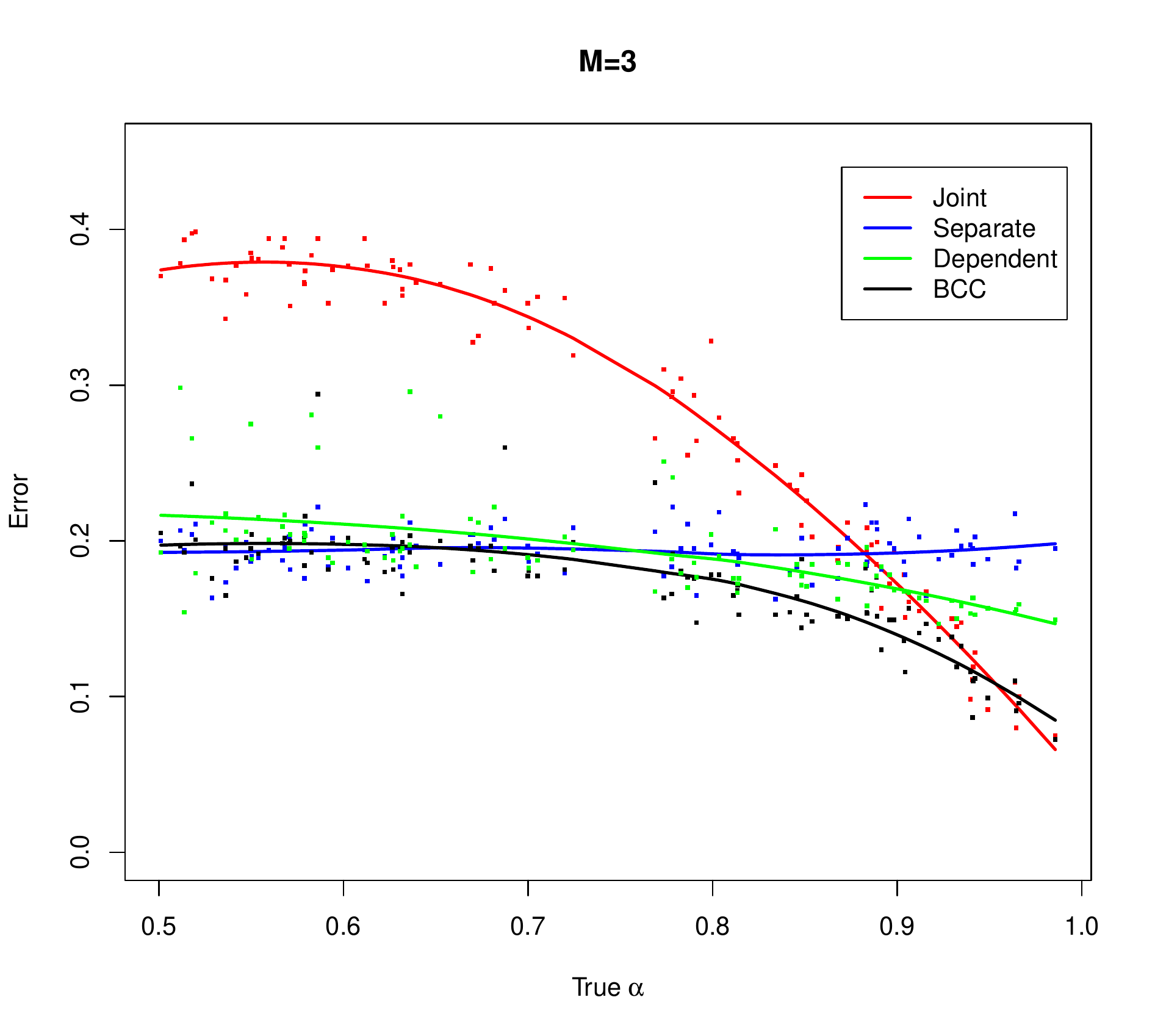}}
\caption{Relative clustering error for 100 simulations with $M=2$ (a) and $M=3$ (b) data sources, shown for separate clustering (\textcolor{blue}{blue}), joint clustering (\textcolor{red}{red}), dependent clustering (\textcolor{green}{green}) and BCC (\textbf{black}).  A LOESS curve displays clustering error as a function of $\alpha$ for each method.   }
\label{fig:Comparison}
\end{center}
\vskip -0.25in
\end{figure}

We consider the relative error for each model in terms of the average number of incorrect cluster assignments: 
\[ \frac{\sum_{m=1}^M \sum_{n=1}^N \mathbbm{1}\{\hat{L}_{mn} \neq L_{mn}\}}{MN}, \]
where $\mathbbm{1}$ is the indicator function.  Note that for joint clustering $\hat{\mathbb{L}}_1 = \hat{\mathbb{L}}_2$.  The relative error for each clustering method on $M=3$ datasets is shown in Figure~\ref{fig:Comparison} (top panel).  Smooth curves are fit to the data using LOESS \cite{Cleveland} and display the relative clustering error for each method as a function of $\alpha$.  Not surprisingly, joint clustering performs well for $\alpha \approx 1$ (perfect agreement) and separate clustering is superior when $\alpha \approx 0.5$ (no relationship). BCC and dependent clustering learn  the level of cluster agreement, and hence serve as a flexible bridge between these two extremes.

The bottom panel of Figure~\ref{fig:Comparison} displays the results for a similar simulation in which $M=3$ data sources are generated.  Again, BCC is competitive with separate clustering when $\alpha \approx 0.5$ and joint clustering when  $\alpha \approx 1$.  Dependent clustering does not perform as well in this example.  The pairwise-dependence model does not assume an overall clustering and hence it has less power to learn the underlying structure for $M>2$.         

\section{Application to Genomic Data}
\label{app}
We apply BCC to multi-source genomic data on breast cancer tumor samples from TCGA.  For a common set of 348 tumor samples, our full dataset includes
\begin{itemize}
\item RNA expression (GE) data for 645 genes.
\item DNA methylation (ME) data for 574 probes.
\item miRNA expression (miRNA) data for 423 miRNAs.
\item Reverse Phase Protein Array (RPPA) data for 171 proteins. 
\end{itemize}
These four data sources are measured on different platforms and represent different biological components.  However, they all represent genomic data for the same samples and it is reasonable to expect some shared structure.  These data are publicly available from the TCGA Data Portal.  See \url{http://people.duke.edu/~el113/software.html} for R code  to completely reproduce the analysis, including instructions on how to download and process these data from the TCGA Data Portal.     

Previously, four comprehensive sample subtypes were identified based on a multi-source consensus clustering of these data \cite{Perou}.  These subtypes were shown to be clinically relevant, as they may be used for more targeted therapies and prognoses. 

We select $K=3$ clusters for BCC, based on the heuristic described in Section~\ref{chooseK}.  Full computational details, as well as charts that illustrate mixing over the MCMC draws, are available in Appendix~\ref{Sec6}.  Table~\ref{tab3} shows a matching matrix comparing the overall clustering $\mathbb{C}$ with the comprehensive subtypes defined by TCGA.  The two partitions have a significant but weak association, and this suggests they may not be driven by the same biological signal.  Similar tables that compare the source-specific clusterings $L_{mn}$ with source-specific subtypes are available in Appendix~\ref{CompareTables}. 

\begin{table}[!ht]
\begin{center}
\begin{tabular}{c c| c c c}
 & & \multicolumn{3}{c}{BCC cluster} \\

& & \textbf{1} & \textbf{2}  & \textbf{3} \\
\hline
 \multirow{4}{*}{TCGA subtype} & \textbf{1} & 13 & 6 & 20\\
 &  \textbf{2} & 66 & 2 & 4\\
 &  \textbf{3} & 3 & 80 & 78\\
 &  \textbf{4} & 0 & 3 & 73\\
\end{tabular}
\caption{BCC cluster vs. TCGA comprehensive subtype matching matrix.  }
\label{tab3}
\end{center}
\end{table}

\begin{figure*}[!ht]
\vskip 0.2in
\begin{center}
\includegraphics[width=\textwidth]{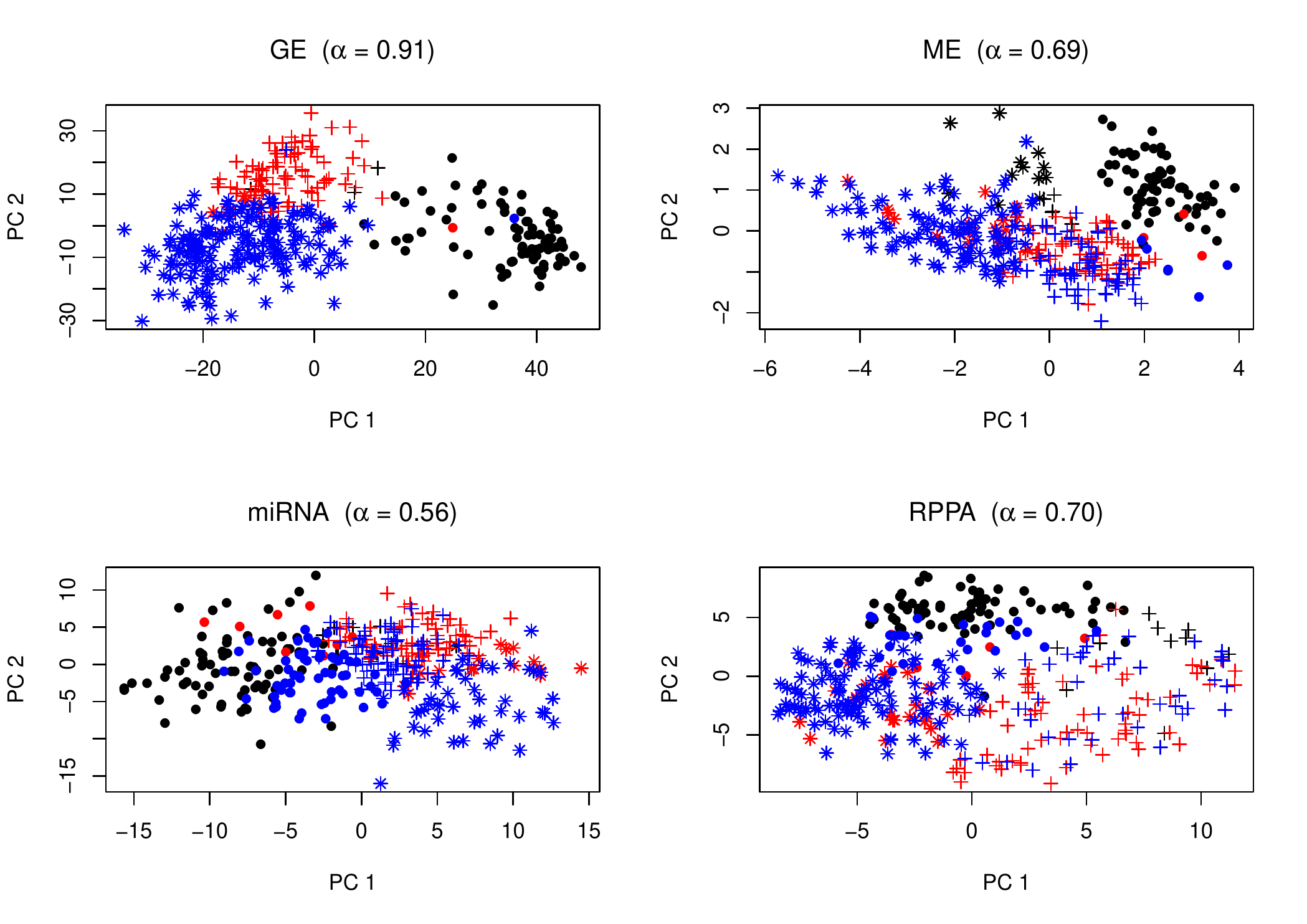}
\caption{PCA plots for each data source. Sample points are colored by overall cluster; cluster 1 is \textbf{black}, cluster 2 is \textcolor{red}{red}, and cluster 3 is \textcolor{blue}{blue}.  Symbols indicate source-specific cluster;  cluster 1 is `$\bullet$', cluster 2 is `$+$', and cluster 3 is `$\ast$' . }
\label{fig:PCA}
\end{center}
\vskip -0.2in
\end{figure*}

Figure~\ref{fig:PCA} provides a point-cloud view of each dataset given by a scatter plot of the the first two principal components.  The overall and source-specific cluster index is shown for each sample, as well as a point estimate for the adherence parameter $\alpha$ (alternatively, clustering heatmaps for each data source are given in Appendix~\ref{AppHeatmaps}).  The four data sources show different structure, and the source-specific clusters are more well distringuished than the overall clusters in each plot.  However, the overall clusters are clearly represented to some degree in all four plots.  Hence, the flexible yet  integrative approach of BCC seems justified for these data.

\section{Conclusions and discussion}

This work was motivated by the perceived need for a general, flexible, and computationally scalable approach to clustering multi-source data. We propose BCC, which models both an overall clustering and a clustering specific to each data source.  We view BCC as a form of consensus clustering, with advantages over traditional methods in terms of modeling uncertainty and the ability to borrow information across sources.  

The BCC model assumes a very simple and general dependence between data sources.  When an overall clustering is not sought, or when such a clustering does not make sense as an assumption, a more general model of cluster dependence (such as MDI) may be more appropriate.  Furthermore, a context-specific approach may be necessary when more is known about the underlying dependence of the data.   For example, \cite{NguyenGelfand} exploit functional covariance models for time-course data to determine overall and time-specific clusters.  

Our implementation of BCC assumes the data are normally distributed and models cluster-specific mean and variance parameters.  It is straightforward to extend this approach to more complex clustering models.  In particular, models that assume clusters exist on a sparse feature set \cite{Tadesse} or allow for more general covariance structure \cite{Ghahramani} are growing in popularity.  While we focus on multi-source biomedical data, the applications of BCC are potentially widespread.  In addition to multi-source data, BCC may be used to compare clusterings from different statistical models for a single homogeneous dataset.

% Acknowledgements should only appear in the accepted version. 
\section*{Acknowledgments} 
This work was supported by grant R01-ES017436 from the National Institute of
Environmental Health Sciences (NIEHS).  
 
%\textbf{Do not} include acknowledgements in the initial version of
%the paper submitted for blind review.

%If a paper is accepted, the final camera-ready version can (and
%probably should) include acknowledgements. In this case, please
%place such acknowledgements in an unnumbered section at the
%end of the paper. Typically, this will include thanks to reviewers
%who gave useful comments, to colleagues who contributed to the ideas, 
%and to funding agencies and corporate sponsors that provided financial 
%support.  

% In the unusual situation where you want a paper to appear in the
% references without citing it in the main text, use \nocite
%\nocite{langley00}

\appendix
%This appendix provides additional details and validation for Bayesian consensus clustering (BCC).  Section~\ref{Sec1} gives full %computational details for the algorithm under a normal-gamma model with cluster-specific mean and variance.  Section~\ref{Sec2} %illustrates the relationship between overall and source-specific cluster sizes.  Section~\ref{Sec3} shows the equivalence of BCC %and MDI, under certain assumptions, when $M=2$.  Section~\ref{Sec4} illustrates the effect of the prior for $\alpha$.  %Section~\ref{Sec5} provides full details for the clustering comparison simulation in the main article.  Section~\ref{Sec6} provides %computational details and further analysis for the application to TCGA data given in the main article.       

\section{Computational details}
\label{Sec1}
\subsection{Normal-gamma model}
\label{NormGam}
Here we fill in the details for a specific case of the Bayesian computational framework given in Section\ref{Comp}.  We assume $\mathbb{X}_i$ has a normal-gamma mixture distribution with cluster-specific mean and variance.  That is,
\[X_{mn} | L_{mn}=k \sim N(\mu_{mk},\Sigma_{mk}),\]
where
\begin{itemize}
\item $\mu_{mk}$ is a $D_m$ dimensional mean vector, where $D_m$ is the dimension of data source $m$.  
\item $\Sigma_{mk}$ is a $D_m \times D_m$ diagonal covariance matrix, $\Sigma_{mk} = \text{Diag}(\sigma_{mk1}, \hdots,\sigma_{mkD_m})$. 
\end{itemize}
We use a $D_m$ dimensional normal-inverse-gamma prior distribution for $\theta_{mk} =  (\mu_{mk}, \Sigma_{mk})$ .  That is, \[\theta_{mk} \sim N \Gamma^{-1} (\eta_{m0},\lambda_0,A_{m0},B_{m0}),\]
where $\eta_{m0}$, $\lambda_0$,$A_{m0}$ and $B_{m0}$ are hyperparameters.  It follows that $\mu_{mk}$ and $\Sigma_{mk}$ are given by
\begin{itemize}
\item $\frac{1}{\sigma_{mkd}^2} \sim \text{Gamma}(A_{m0d},B_{m0d})$, and
\item $\mu_{mkd} \sim N(\eta_{m0},\frac{\sigma_{mkd}^2}{\lambda_0})$ for $d=1,\hdots, D_m.$
\end{itemize}
By default we set $\lambda_0 =1$, and estimate $\mu_{m0},A_{m0}$ and $B_{m0}$ from the mean and variance of each variable in $\mathbb{X}_m$.  

The $i$'th iteration in the MCMC sampling scheme procedes as follows:  
\begin{enumerate}
\item Generate $\Theta_m^{(i)}$ given $\{\mathbb{X}_m,\mathbb{L}_m^{(i-1)}\}$, for $m=1,\hdots,M$. The posterior distribution for $\theta_{mk}^{(i)}$, $k=1,\hdots,K$ is
\[\theta_{mk}^{(i)} \sim N \Gamma^{-1} (\eta_{mk}^{(i)}, \lambda_k^{(i)},A_{m0}^{(i)},B_{m0}^{(i)}).\]
Let $N_{mk}$ be the number of samples allocated to cluster $k$ in $\mathbb{L}_m^{(i-1)}$, $\bar{X}_{mk}$ be the sample mean vector for cluster $k$, and $S_{mk}$ the sample variance vector for cluster $k$. The posterior normal-gamma parameters are 
\begin{itemize}
\item $\eta_{mk}^{(i)} = \frac{\lambda_0 \eta_{m0} + N_{mk} \bar{X}_{mk}}{\lambda_0+N_{mk}}$ 
\item  $\lambda_k^{(i)} = \lambda_0+N_{mk}$ 
\item $A_{m0}^{(i)}=A_{m0}+\frac{n}{2}$
\item $B_{m0}^{(i)} = B_{m0}+\frac{N_{mk}S_{mk}}{2}+\frac{\lambda_0 N_{mk}(\bar{X}_{mk}-\mu_{m0})^2}{2(\lambda_0+N_{mk})} $.
\end{itemize}  

\item Generate $\mathbb{L}_{m}^{(i)}$  given $\{\mathbb{X}_m,\Theta_m^{(i)},\alpha_m^{(i-1)}, \mathbb{C}^{(i-1)}\}$, for $m=1,\hdots,M$.  The posterior probability that $L_{mn}^{(i)}=k$ for $k=1,\hdots,K$ is proportional to 
\[\nu(k,C_n^{(i-1)},\alpha_m^{(i-1)})f_m(X_{mn}|\theta_{mk}^{(i)}),\]
where $f_m$ is the multivariate normal density defined by $\theta_{mk} = (\mu_{mk},\Sigma_{mk})$.

\item Generate $\alpha_m^{(i)}$ given $\{\mathbb{C}^{(i-1)},\mathbb{L}_m^{(i)}\}$, for $m=1,\hdots,M$.  The posterior distribution for $\alpha_m^{(i)}$ is TBeta$(a_m+\tau_m,b_m+N-\tau_m,\frac{1}{K})$, where $\tau_m$ is the number of samples $n$ satisfying $L_{mn}^{(i)} = C_{n}^{(i-1)}$.

\item Generate $\mathbb{C}^{(i)}$ given $\{\mathbb{L}_m^{(i)},\Pi^{(i-1)},\alpha^{(i)}\}$.  The posterior probability that $C_n^{(i)}=k$ for $k=1,\hdots,K$ is proportional to 
\[\pi_k \prod_{m=1}^M \nu(k,L_{mn}^{(i)},\alpha_m^{(i)}) .\] 

\item Generate $\Pi^{(i)}$ given $\mathbb{C}^{(i)}$.  The posterior distribution for $\Pi^{(i)}$ is Dirichlet $(\beta_0+\rho)$ where $\rho_k$ is the number of samples allocated to cluster $k$ in $\mathbb{C}^{(i)}$.  
\end{enumerate}

By default, we initialize $\mathbb{L}_1,\hdots,\mathbb{L}_k$ by a K-means clustering of each dataset.  After running the Markov chain for a specified number of iterations (e.g., 10000), the method described in \cite{Dahl} is used to determine a hard clustering for each of $\mathbb{C}$, $\mathbb{L}_1,\hdots,\mathbb{L}_m$.   

\subsection{Assuming equal adherence}  \label{appendix1.2}
It is straightforward to modify the procedure above under the assumption that each data source adheres equally well to the overall clustering $\mathbb{C}$.  Rather than modeling $\alpha_1,\hdots,\alpha_m$ separately, we assume $\alpha = \alpha_1=\hdots=\alpha_m$.  The prior for $\alpha$ is a truncated beta distribution:
\[\alpha \sim \text{TBeta}(a,b,\frac{1}{K}).\] 
The MCMC sampling scheme procedes exactly as above, except that in step (3) we need only generate $\alpha^{(i)}$ from the posterior distribution TBeta$(a+\tau, b + NM-\tau,\frac{1}{K})$.  Here $\tau=\sum_{m=1}^M \tau_m$, where the $\tau_m$ are as defined in step (3) above.  

\section{Cluster size illustration}
\label{Sec2}
Recall that $\pi_k$ is the marginal probability that an object belongs to the overall cluster $k$: 
\[\pi_k = P(C_n=k).\] The probability that an object belongs to a given source-specific cluster is then
\[P(L_{mn}=k|\Pi)= \pi_k \alpha_m+(1-\pi_k)\frac{1-\alpha_m}{K-1}.\]  
As a consequence, the size of the source-specific clusters are generally more uniform than the size of the overall clusters.  In particular, the source-specific clusterings $\mathbb{L}_m$ will generally represent more clusters than $\mathbb{C}$, rather than vice-versa.    

\begin{figure*}[!ht]
\vskip 0.2in
\begin{center}
\includegraphics[width=0.9\textwidth, trim = 0mm 5mm 0mm 5mm, clip = TRUE]{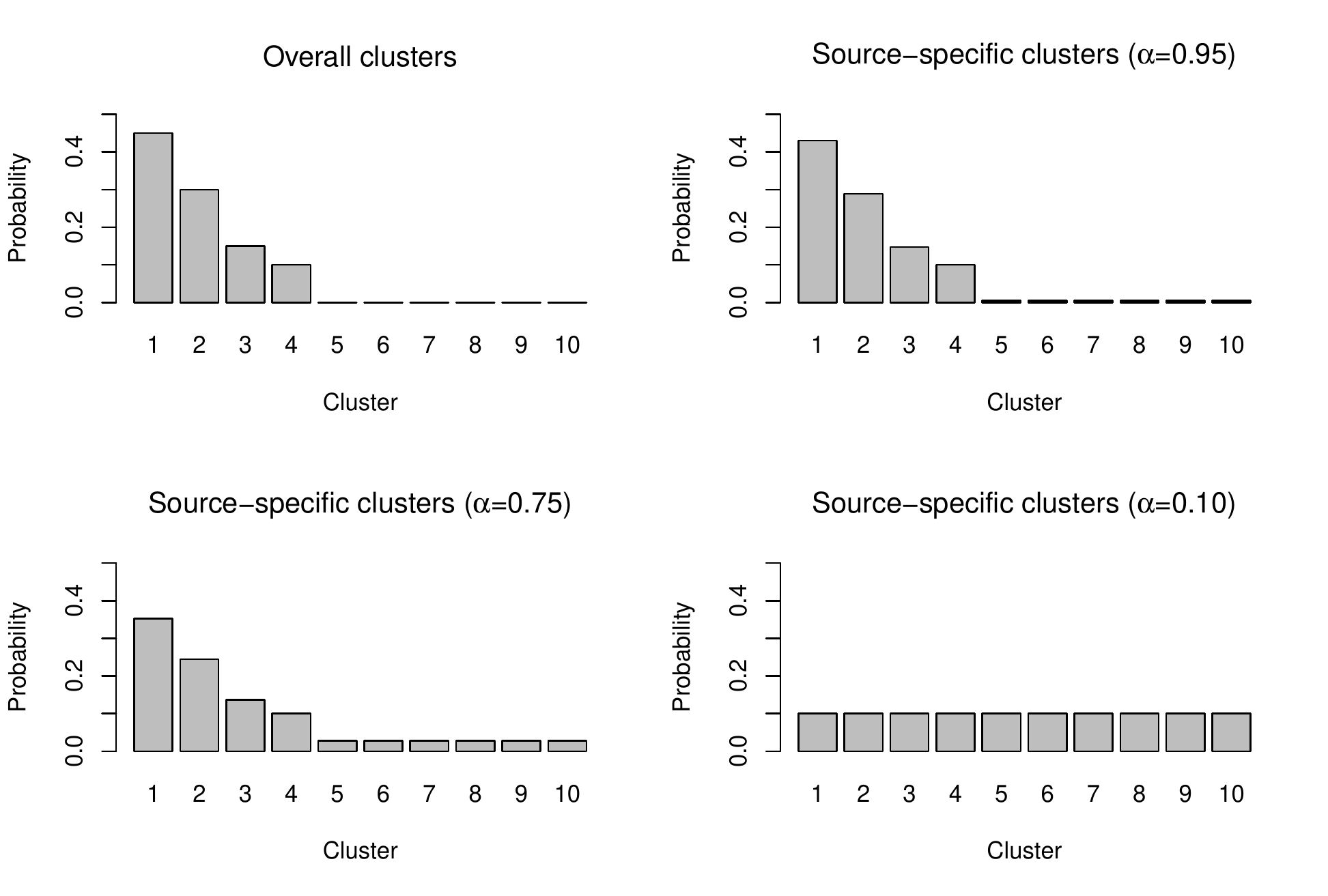}
\caption{Marginal cluster inclusion probabilities are shown for the overall clusters ($\Pi$) with $K=10$ (top-left).  The source-specific cluster probabilities induced by $\Pi$ are shown for the adherence levels $\alpha_m = 0.95, 0.75$ and $0.10$. } \label{fig:ClusterProbs}
\end{center}
\vskip -0.2in
\end{figure*}  

As an illustration, we set $K=10$ and assume the $\tau_k$ have the skewed distribution shown in the top left panel of Figure~\ref{fig:ClusterProbs}.  The marginal cluster inclusion probabilities for a given data source depend on its adherence $\alpha_m$ to the overall clustering.  Not surprisingly, for $\alpha_m$ close to 1 the inclusion probabilities closely resemble those for the overalll clustering (top right panel of Figure~\ref{fig:ClusterProbs}).  As $\alpha_m$ approaches $\frac{1}{K}$, the inclusion probabilities are more uniform (bottom two panels of Figure \ref{fig:ClusterProbs}).  In particular, clusters that had zero probability to occur in $\mathbb{C}$ have positive probability to occur in $\mathbb{L}_m$.  Hence, a sample that does not fit any overall pattern in a given data source (e.g., an outlier) need not be allocated to an overall cluster.

\section{Equivalence of BCC and MDI}
\label{Sec3}
Here we compare the MDI and BCC models for $M=2$ data sources, giving conditions where the two models are equivalent under a parameter substitution.  

Assume $\alpha=\alpha_1=\alpha_2$, and let $U=\frac{\alpha}{1-\alpha}$.  The joint distribution of $(\mathbb{L}_1,\mathbb{L}_2)$ under BCC is then  
\begin{eqnarray*}P(\{L_{mn}=k_m\}|\Pi,\mathbb{\alpha}) \propto \left\{\hspace{-5 pt} \begin{array}{ccl} \pi_1 U^2+(1-\pi_1) &\hspace{-5 pt}\text{if}&\hspace{-5 pt} k_1=k_2=1 \\ \pi_1+ (1-\pi_1)U^2 &\hspace{-5 pt}\text{if}&\hspace{-5 pt} k_1=k_2=2 \\  U &\hspace{-5 pt}\text{if}& \hspace{-5 pt} k_1 \neq k_2 . \end{array} \right.\end{eqnarray*}
Assume $\tilde{\pi}= \tilde{\pi}_{1 \cdot} = \tilde{\pi}_{2 \cdot}$ and let $\phi=\phi_{12}$ in the MDI clustering model.  The joint distribution of $(\mathbb{L}_1,\mathbb{L}_2)$ under MDI is then
\begin{eqnarray*}P(\{L_{mn}=k_m\}|\tilde{\Pi},\phi) \propto \left\{\hspace{-5 pt} \begin{array}{ccl} \tilde{\pi}_1^2 (1+\phi) &\hspace{-5 pt}\text{if}&\hspace{-5 pt} k_1=k_2=1 \\ (1-\tilde{\pi}_1)^2(1+\phi) &\hspace{-5 pt}\text{if}&\hspace{-5 pt} k_1=k_2=2 \\  \tilde{\pi}_1(1-\tilde{\pi}_1) &\hspace{-5 pt}\text{if}& \hspace{-5 pt} k_1 \neq k_2 . \end{array} \right.\end{eqnarray*}
It is straightforward to verify that the two forms are equivalent under the substitutions 
\[\phi = \sqrt{\left((1-\pi_1)U+\frac{\pi_1}{U} \right) \left( \pi_1 U+\frac{1-\pi_1}{U} \right)}-1\]
and
\[\tilde{\pi}_1 = \frac{\sqrt{(1-\pi_1)U^{-1}+\pi_1U}}{\sqrt{(1-\pi_1)U+\pi_1 U^{-1}}+\sqrt{(1-\pi_1)U^{-1}+\pi_1U}} .\]

There is no such equivalence for $M>2$, regardless of restrictions on $\Pi$ and $\Phi$.  

\section{Prior comparison for $\alpha$}
\label{Sec4}
We use a simple simulation to illustrate the effect of the prior distribution for $\alpha$.  We generate  datasets $\mathbb{X}_1: 1 \times 200$ and $\mathbb{X}_2:1 \times 200$ as in Section\ref{AlphaAcc}:
\begin{enumerate}
\item Let $\mathbb{C}$ define two clusters, where $C_n = 1$ for $n \in \{1,\hdots,100\}$ and $C_n = 2$ for $n \in \{101,\hdots,200\}$ 
\item Draw $\alpha$ from a Uniform$(0.5,1)$ distribution.
\item For $m=1,2$ and $n=1,\hdots,200$, generate $L_{mn} \in \{1,2\}$ so that $P(L_{mn} = C_n) = \alpha$ and $P(L_{mn} \neq C_n) = 1-\alpha$.
\item For $m=1,2$ draw values $X_{mn}$ from a Normal$(1.5,1)$ distribution if $L_{mn}=1$ and from a Normal$(-1.5,1)$ distribution if $L_{mn}=2$ 
\end{enumerate}

\begin{figure*}[!ht]
\vskip 0.2in
\begin{center}
\includegraphics[width=0.9\textwidth, trim = 0mm 0mm 0mm 0mm, clip = TRUE]{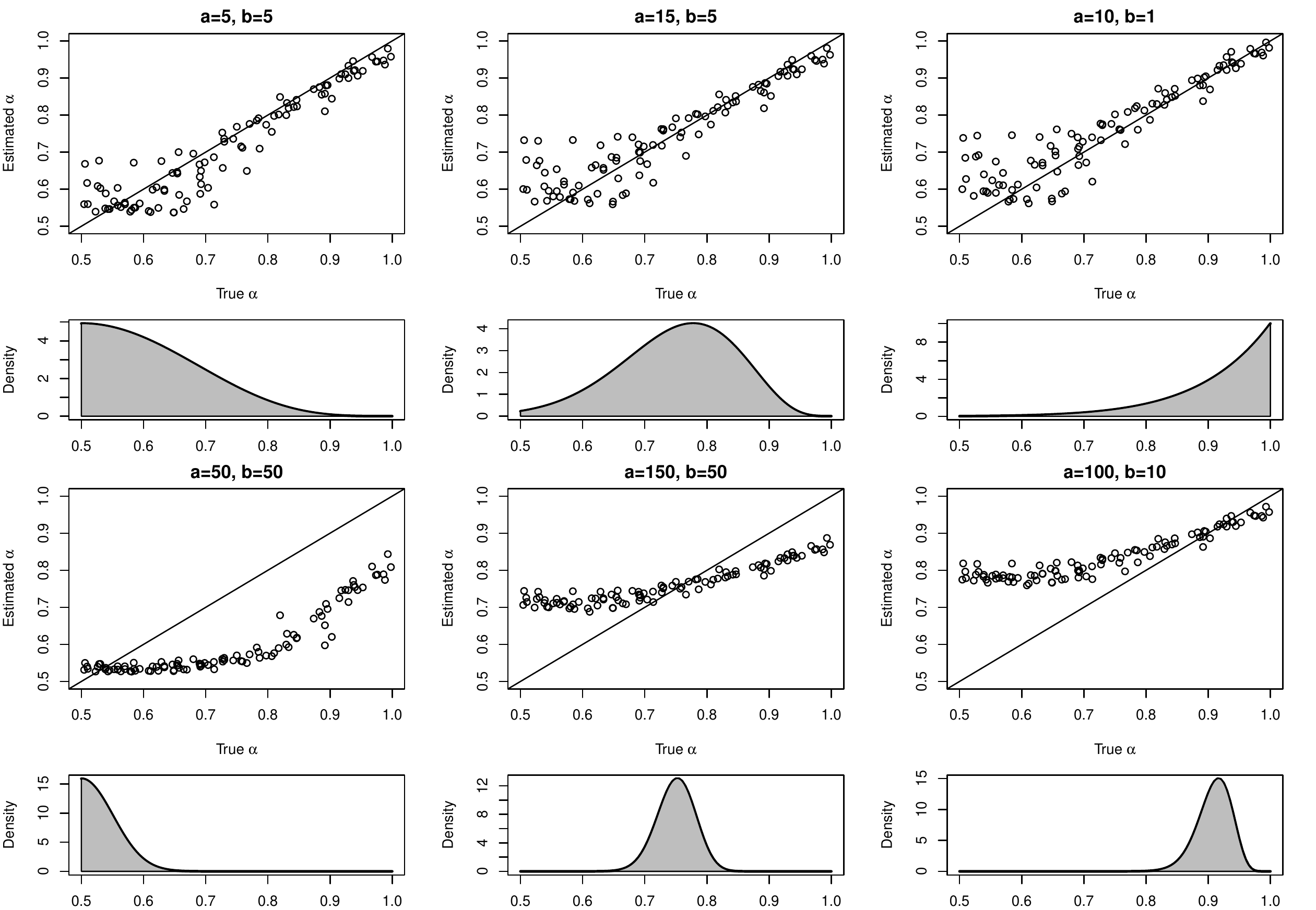}
\caption{Scatterplots of $\hat{\alpha}$ versus the true $\alpha$ are shown for various prior distributions on $\alpha$.  Each prior is of the form TBeta$(a,b,\frac{1}{2})$, and the density of each prior is shown below the relevant scatterplot.}
\label{fig:AlphaPriors}
\end{center}
\vskip -0.2in
\end{figure*}

We estimate the BCC model under the assumption that $\alpha=\alpha_1=\alpha_2$, where $\alpha$ has prior distribution TBeta$(a,b,\frac{1}{2})$, for various values of $a$ and $b$.  The uniform prior ($a=b=1$) gives relatively unbiased results, as illustrated in Figure~\ref{fig:alphaEsts} of the main mansuscript.  Figure~\ref{fig:AlphaPriors} displays the estimated values $\hat{\alpha}$ for alternative choices of $a$ and $b$.  Not surprisingly, for very precise priors (large $a$ and $b$) the $\hat{\alpha}$ are highly influenced by the prior and are therefore inaccurate.  However, the $\hat{\alpha}$ appear to be robust for moderately precise priors.     

\section{Clustering comparison details}
\label{Sec5}
Here we describe the computational details for the four procedures used in the clustering comparison study given in Section~\ref{ClusterComp}.  For each procedure the MCMC algorithm ran for 1000 iterations (after 200 iterations of ``burn-in"), and a hard clustering was determined as in \citet{Dahl}.  
\subsection{Separate clustering}
\label{SepClust}
We use a normal-gamma mixture model to cluster each $\mathbb{X}_m$.  The marginal probability that $X_{mn}$ is allocated to cluster $k$ is $\pi_{mk}$, where $\Pi_m= (\pi_{m1},\hdots,\pi_{mk}) \sim \text{Dirichlet}(\beta_0)$.  We use $K=2$ clusters and $\beta_0 = (1,1)$.  The $i$'th iteration in the MCMC sampling scheme procedes as follows:  
\begin{enumerate}
\item Generate $\Theta_m^{(i)}$ given $\{\mathbb{X}_m,\mathbb{L}_m^{(i-1)}\}$. The posterior distribution for $\theta_{mk}^{(i)}$, $k=1,\hdots,K$ is
\[\theta_{mk}^{(i)} \sim N \Gamma^{-1} (\eta_{mk}^{(i)}, \lambda_k^{(i)},A_{m0}^{(i)},B_{m0}^{(i)}).\]
Let $N_{mk}$ be the number of samples allocated to cluster $k$ in $\mathbb{L}_m^{(i-1)}$, $\bar{X}_{mk}$ be the sample mean vector for cluster $k$, and $S_{mk}$ the sample variance vector for cluster $k$. The posterior normal-gamma parameters are 
\begin{itemize}
\item $\eta_{mk}^{(i)} = \frac{\lambda_0 \eta_{m0} + N_{mk} \bar{X}_{mk}}{\lambda_0+N_{mk}}$ 
\item  $\lambda_k^{(i)} = \lambda_0+N_{mk}$ 
\item $A_{m0}^{(i)}=A_{m0}+\frac{n}{2}$
\item $B_{m0}^{(i)} = B_{m0}+\frac{N_{mk}S_{mk}}{2}+\frac{\lambda_0 N_{mk}(\bar{X}_{mk}-\mu_{m0})^2}{2(\lambda_0+N_{mk})} $.
\end{itemize}  

\item Generate $\mathbb{L}_{m}^{(i)}$  given $\{\mathbb{X}_m,\Theta_m^{(i)},\Pi_m^{(i-1)}\}$.  The posterior probability that $L_{mn}^{(i)}=k$ for $k=1,\hdots,K$ is proportional to 
\[\pi_{mk}f_m(X_{mn}|\theta_{mk}^{(i)}),\]
where $f_m$ is the multivariate normal density defined by $\theta_{mk} = (\mu_{mk},\Sigma_{mk})$.

\item Generate $\Pi_m^{(i)}$ given $\mathbb{L}_m^{(i)}$.  The posterior distribution for $\Pi_mi=^{(i)}$ is Dirichlet $(\beta_0+\rho_{m})$ where $\rho_{mk}$ is the number of samples allocated to cluster $k$ in $\mathbb{L}_m^{(i)}$.  
\end{enumerate}

\subsection{Joint clustering}
We use a normal-gamma mixture model to cluster the concatenated dataset \[\mathbb{X} = \left[ \begin{array}{c} \mathbb{X}_1 \\ \vdots \\ \mathbb{X}_M \end{array}\right].\]
The computational details are exactly as in Section~\ref{SepClust}, except that we perform the algorithm on the joint data $\mathbb{X}$ rather than separately for each $\mathbb{X}_m$.  

\subsection{Dependent clustering}
For our dependent clustering model we let $\alpha_{m_1m_2}$ be the probability that $L_{m_1n} =L_{m_2n}$, where $\alpha_{m_1m_2} \sim \text{TBeta}(a,b,\frac{1}{K}$). Hence, we model the clustering dependence between each pair of datasets, rather than adherence to an overall clustering. The marginal probability that $X_{mn}$ is allocated to cluster $k$ is $\pi_{mk}$, where $\Pi_m= (\pi_{m1},\hdots,\pi_{mk}) \sim \text{Dirichlet}(\beta_0)$.  We use $K=2$ clusters,$\beta_0 = (1,1)$, and $a=b=1$.  The $i$'th iteration in the MCMC sampling scheme procedes as follows:  
\begin{enumerate}
\item Generate $\Theta_1^{(i)}$ given $\{\mathbb{X}_m,\mathbb{L}_m^{(i-1)}\}$, as in step (1) of Section~\ref{NormGam}. 

\item Generate $\mathbb{L}_{1}^{(i)},\hdots,\mathbb{L}_{M}^{(i)}$ given $\{\mathbb{X},\mathbbm{\alpha}^{(i-1)},\Theta^{(i)},\Pi^{(i-1)}\}$. For $n=1,\hdots,N$, the joint posterior probability for $\{L_{mn}=k_m\}_{m=1}^M$ is proportional to 
\[\prod_{m=1}^{M} \pi_{mk}^{(i-1)} f_m(X_{mn}|\theta_{mk}^{(i)}) \prod_{m'=m+1}^M \nu(k_{m},k_{m'},\alpha_{mm'}^{(i-1)}).\]

\item Generate $\alpha_{m_1m_2}^{(i)}$ given $\{L_{m_1}^{(i)},L_{m_2}^{(i)}\}$, for all pairs $\{(m_1,m_2): m_1 = 1,\hdots,M \text{ and } m_2= (m_1~+~1),\hdots,M\}$ .    The posterior distribution for $\alpha_{m_1m_2}^{(i)}$ is TBeta$(a+\tau_{m_1m_2},b+N-\tau_{m_1m_2},\frac{1}{K})$, where $\tau_{m_1m_2}$ is the number of samples $n$ satisfying $L_{m_1n}^{(i)} = L_{m_2n}^{(i)}$.  

\item  Generate $\Pi_m^{(i)}$ given $\mathbb{L}_m^{(i)}$.  The posterior distribution for $\Pi_m^{(i)}$ is Dirichlet $(\beta_0+\rho_{m})$ where $\rho_{mk}$ is the number of samples allocated to cluster $k$ in $\mathbb{L}_m^{(i)}$. 
 
 \end{enumerate}

\subsection{BCC}
We implement BCC as described in Section~\ref{Sec1}, assuming equal adherence.  We use $K=2$ clusters, $\beta_0 = (1,1)$, and $a=b=1$. 

\section{TCGA application details}
\label{Sec6}
Here we provide more details for the application to heterogenous genomic data from TCGA that is presented in Section~\ref{app}.  We focus on the application of BCC to GE, ME, miRNA and RPPA data for 348 breast cancer tumor samples.  For more information on the origin of these data and pre-processing details see \cite{Perou}.  

\subsection{Choice of $K$}
\begin{figure}[!ht]
\vskip 0.2in
\begin{center}
\centerline{\includegraphics[width=\columnwidth, trim = 0mm 5mm 0mm 7mm, clip = TRUE]{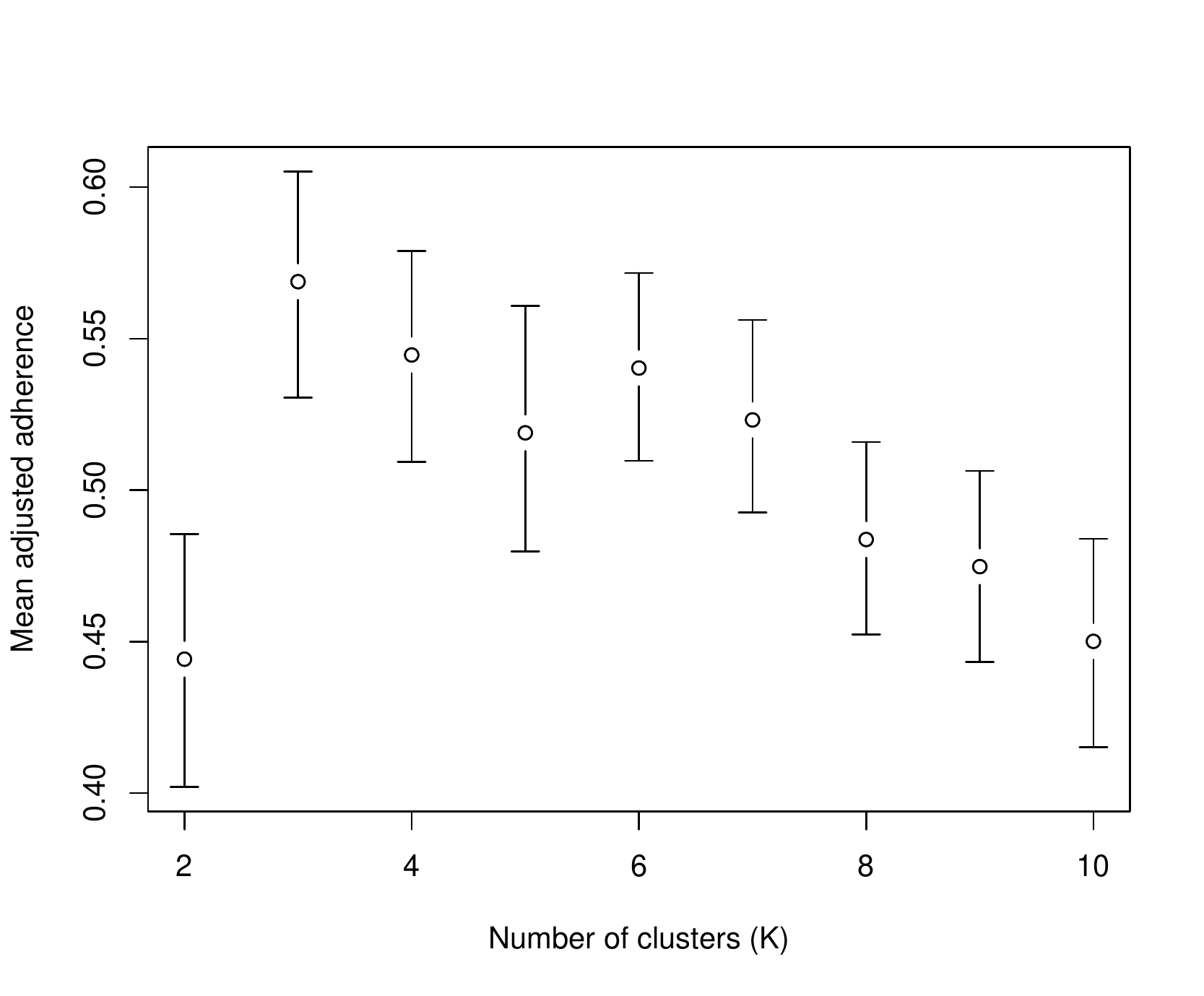}}
\caption{The mean adjusted adherence $\bar{\alpha}_K^*$ is shown after estimating the model with $K=2,\hdots,10$.  A point estimate given by the average of the MCMC draws, and a credible interval given by the  $2.5$ and $97.5$ percentiles of the MCMC draws, are shown for each $K$.}
\label{fig:Kadherence}
\end{center}
\vskip -0.2in
\end{figure}

We use the heuristic method described in Section 4.1 of the main article to choose the number of clusters $K$.  The full BCC model is estimated as in Section~\ref{Sec1}, for the potential values $K=2,\hdots,10$.  We model the $\alpha_m$ separately, set $a_m=b_m=1$  for each $m$, and set $\beta_0 = (1,\hdots,1)$.  We run the MCMC sampling scheme for $10,000$ iterations for each $K$ (the first $2000$ iterations are used as a ``burn-in").  We compute the mean adjusted adherence $\bar{\alpha}_K^*$ for each $K$.  Figure~\ref{fig:Kadherence} shows a point estimate for $\bar{\alpha}_K^*$  (an average over the MCMC iterations), as well as a $95\%$ credible interval based on the MCMC iterations, as a function of $K$.  The maximum value is obtained for $K=3$ (          
$\bar{\alpha}_K^*=0.57$), although the adherence level is comparable for $K=4$ and $K=6$.  

\subsection{MCMC mixing}
\begin{figure}[!ht]
\vskip 0.2in
\begin{center}
\centerline{\includegraphics[width=\columnwidth, trim = 0mm 5mm 0mm 7mm, clip = TRUE]{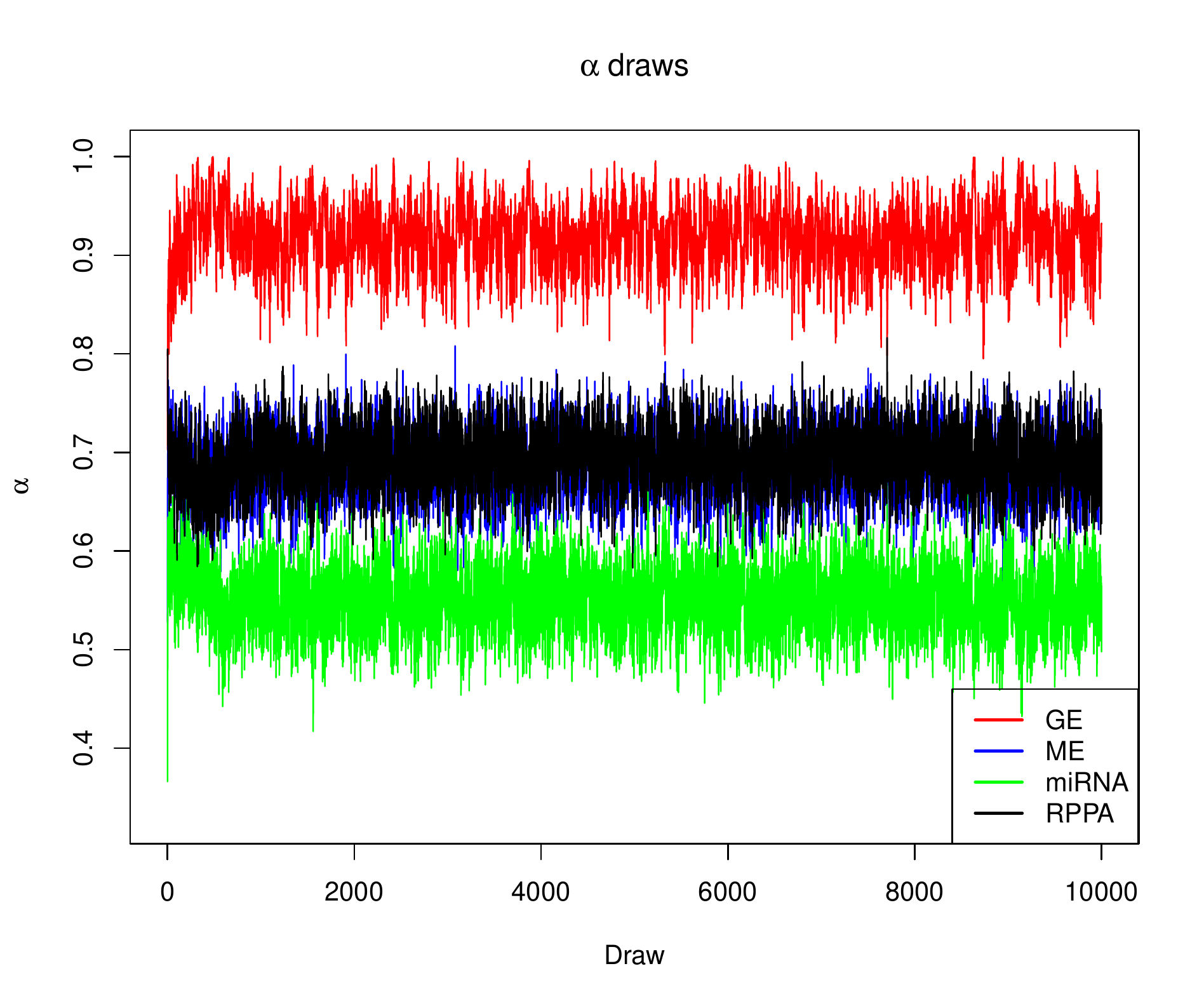}}
\caption{Values of $\alpha_m$ are shown for the data sources GE, ME, miRNA and RPPA over the 10,000 MCMC draws. }
\label{fig:AlphaDraws}
\end{center}
\vskip -0.2in
\end{figure}

\begin{figure}[!ht]
\vskip 0.2in
\begin{center}
\centerline{\includegraphics[width=\columnwidth, trim = 0mm 5mm 0mm 7mm, clip = TRUE]{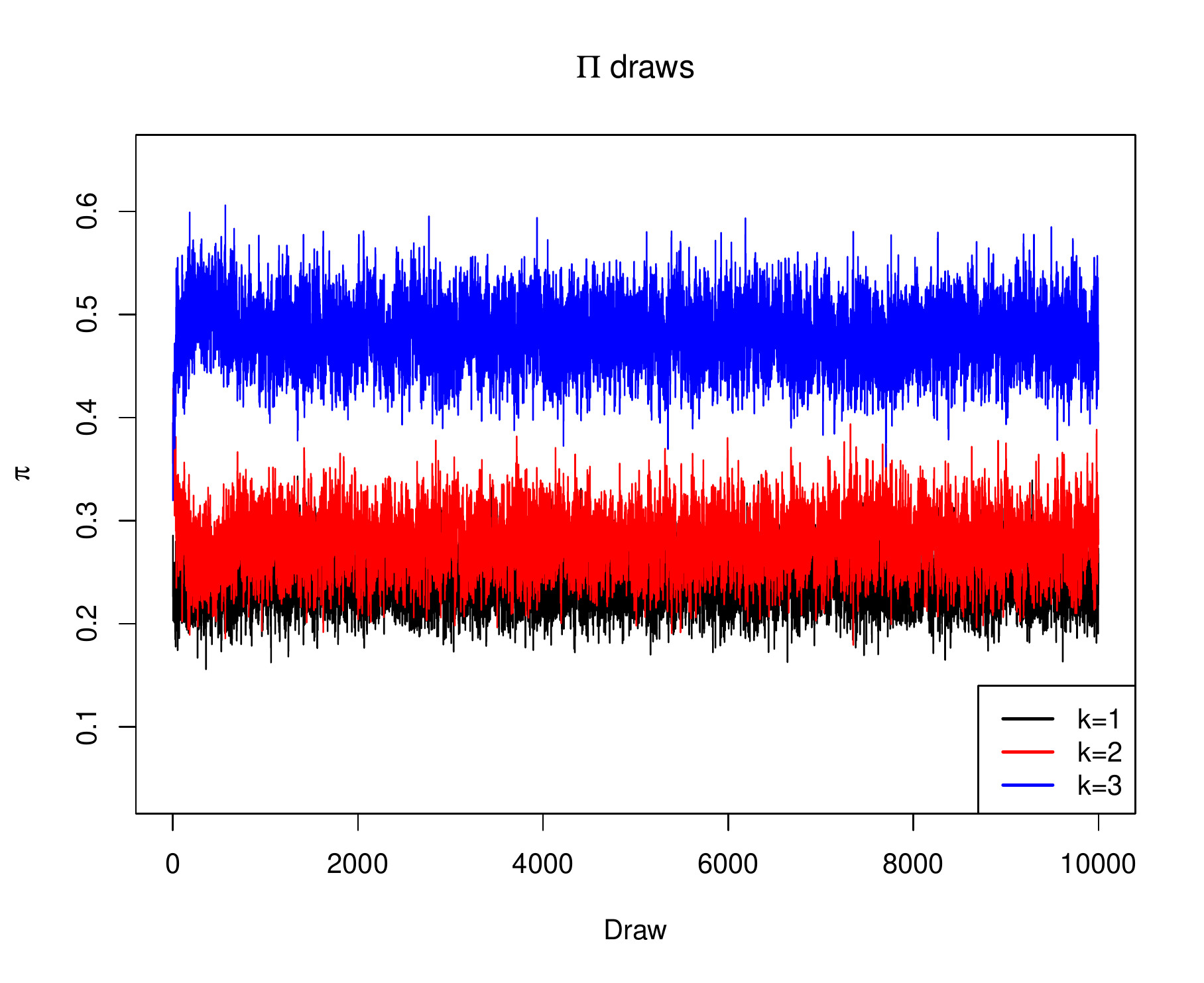}}
\caption{Values of $\pi_k$ are shown for  $k=1,2,3$ over the 10,000 MCMC draws. }
\label{fig:PiDraws}
\end{center}
\vskip -0.2in
\end{figure}

We consider the $10,000$ MCMC draws for $K=3$.  The draws appear to mix well and they converge quickly to a stationary posterior distribution.  Figure~\ref{fig:AlphaDraws} shows the draws for $\alpha_m$, $m=1,\hdots,4$.  These are the estimated adherence to the overall clustering for GE, ME, miRNA and RPPA.  They appear to converge within the first 1000 iterations to an approximately stationary distribution.  The average values are $\alpha = 0.91$ for GE, $\alpha = 0.69$ for ME, $\alpha= 0.56$ for miRNA and $\alpha = 0.70$ for RPPA.  Figure~\ref{fig:PiDraws} shows the marginal overall cluster inclusion probabilities $\pi_k$ over the MCMC draws.  These also quickly converge to a stationary distribution.  The average values are $\hat{\pi}_1 = 0.24$, $\hat{\pi}_2 = 0.28$ and $\hat{\pi}_3 = 0.48$. 

\subsection{Comparison with TCGA subtypes} \label{CompareTables}
Table 3 of the main article compares the overall clusters identified by BCC with the four overall subtypes defined by TCGA.  TCGA has also defined subtypes that are particular to each data source, and we compare the source-specific clusterings identified by BCC with the source-specific subtypes defined by TCGA.  Tables~\ref{tabGE},~\ref{tabME},~\ref{tabmiRNA}, and~\ref{tabRPPA} show the cluster vs. subtype matching matrix for GE, ME, miRNA and RPPA, respectively.  The subtypes for GE and RPPA are named by TCGA according to their biological profile.  In all cases the two sample partitions have a significant association (p-value $<0.01$; Fisher's exact test).  However, these associations are not  exceptionally strong, suggesting that the two partitions may not be driven by the same structure in the data. 
 
\begin{table}[!ht]
\begin{center}
\begin{tabular}{c c| c c c}
 & & \multicolumn{3}{c}{BCC cluster} \\
\multicolumn{2}{c|}{\textbf{GE}} & \textbf{1} & \textbf{2}  & \textbf{3} \\
\hline
 \multirow{5}{*}{TCGA subtype} & \textbf{Basal} & 65 & 1 & 0\\
 &  \textbf{HER2} & 14 & 5 & 23\\
 &  \textbf{Luminal A} & 0 & 78 & 76\\
 &  \textbf{Luminal B} & 0 & 5 & 76\\
 &  \textbf{Normal} & 2 & 3 & 0\\
\end{tabular}
\caption{BCC cluster vs. TCGA subtype matching matrix for GE data.  The GE subtypes are named according to their biological profile.}
\label{tabGE}
\end{center}
\end{table}

\begin{table}[!ht]
\begin{center}
\begin{tabular}{c c| c c c}
 & & \multicolumn{3}{c}{BCC cluster} \\
\textbf{ME}& & \textbf{1} & \textbf{2}  & \textbf{3} \\
\hline
 \multirow{5}{*}{TCGA subtype} & \textbf{1} & 0 & 46 & 12\\
 &  \textbf{2} & 1 & 88 & 1\\
 &  \textbf{3} & 0 & 0 & 36\\
 &  \textbf{4} & 0 & 3 & 87\\
 &  \textbf{5} & 73 & 0 & 1\\
\end{tabular}
\caption{BCC cluster vs. TCGA subtype matching matrix for ME data.  }
\label{tabME}
\end{center}
\end{table}

\begin{table}[!ht]
\begin{center}
\begin{tabular}{c c| c c c}
 & & \multicolumn{3}{c}{BCC cluster} \\
\textbf{miRNA}& & \textbf{1} & \textbf{2}  & \textbf{3} \\
\hline
 \multirow{7}{*}{TCGA subtype} & \textbf{1} & 10 & 9 & 8\\
 &  \textbf{2} & 10 & 14 & 25\\
 &  \textbf{3} & 23 & 4 & 4\\
 &  \textbf{4} & 41 & 66 & 9\\
 &  \textbf{5} & 47 & 3 & 0\\
 &  \textbf{6} & 8 & 43 & 5\\
 &  \textbf{7} & 1 & 8 & 10\\
\end{tabular}
\caption{BCC cluster vs. TCGA subtype matching matrix for miRNA data.  }
\label{tabmiRNA}
\end{center}
\end{table}

\begin{table}[!ht]
\begin{center}
\begin{tabular}{c c| c c c}
 & & \multicolumn{3}{c}{BCC cluster} \\
\multicolumn{2}{c|}{\textbf{RPPA}} & \textbf{1} & \textbf{2}  & \textbf{3} \\
\hline
 \multirow{5}{*}{TCGA subtype} & \textbf{Basal} & 61 & 1 & 7\\
 &  \textbf{Her2} & 29 & 1 & 13\\
 &  \textbf{Luminal A/B} & 1 & 14 & 105\\
 &  \textbf{ReacI} & 0 & 61 & 2\\
 &  \textbf{ReacII} & 7 & 34 & 0\\
\end{tabular}
\caption{BCC cluster vs. TCGA subtype matching matrix for RPPA data.  The RPPA subtypes are named according to their biological profile.}
\label{tabRPPA}
\end{center}
\end{table}

\subsection{Heatmaps}\label{AppHeatmaps}
\begin{figure}[!ht]
\vskip 0.2in
\begin{center}
\centerline{\includegraphics[width=\columnwidth, trim = 0mm 5mm 0mm 7mm, clip = TRUE]{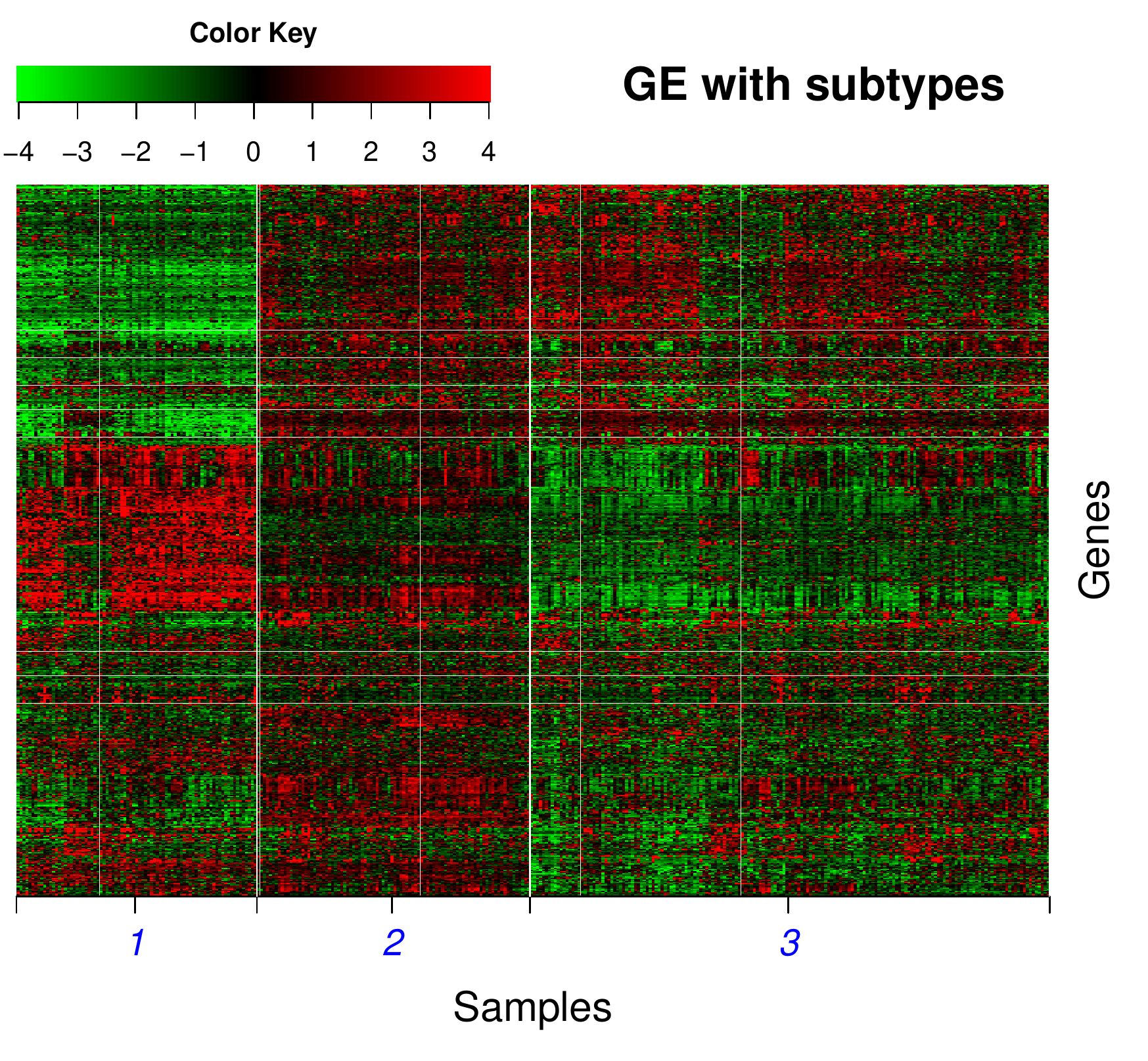}}
\caption{Heatmap of the GE data;  samples are grouped by their GE-specific cluster.}
\label{fig:GEHeatmap}
\end{center}
\vskip -0.2in
\end{figure}

\begin{figure}[!ht]
\vskip 0.2in
\begin{center}
\centerline{\includegraphics[width=\columnwidth, trim = 0mm 5mm 0mm 7mm, clip = TRUE]{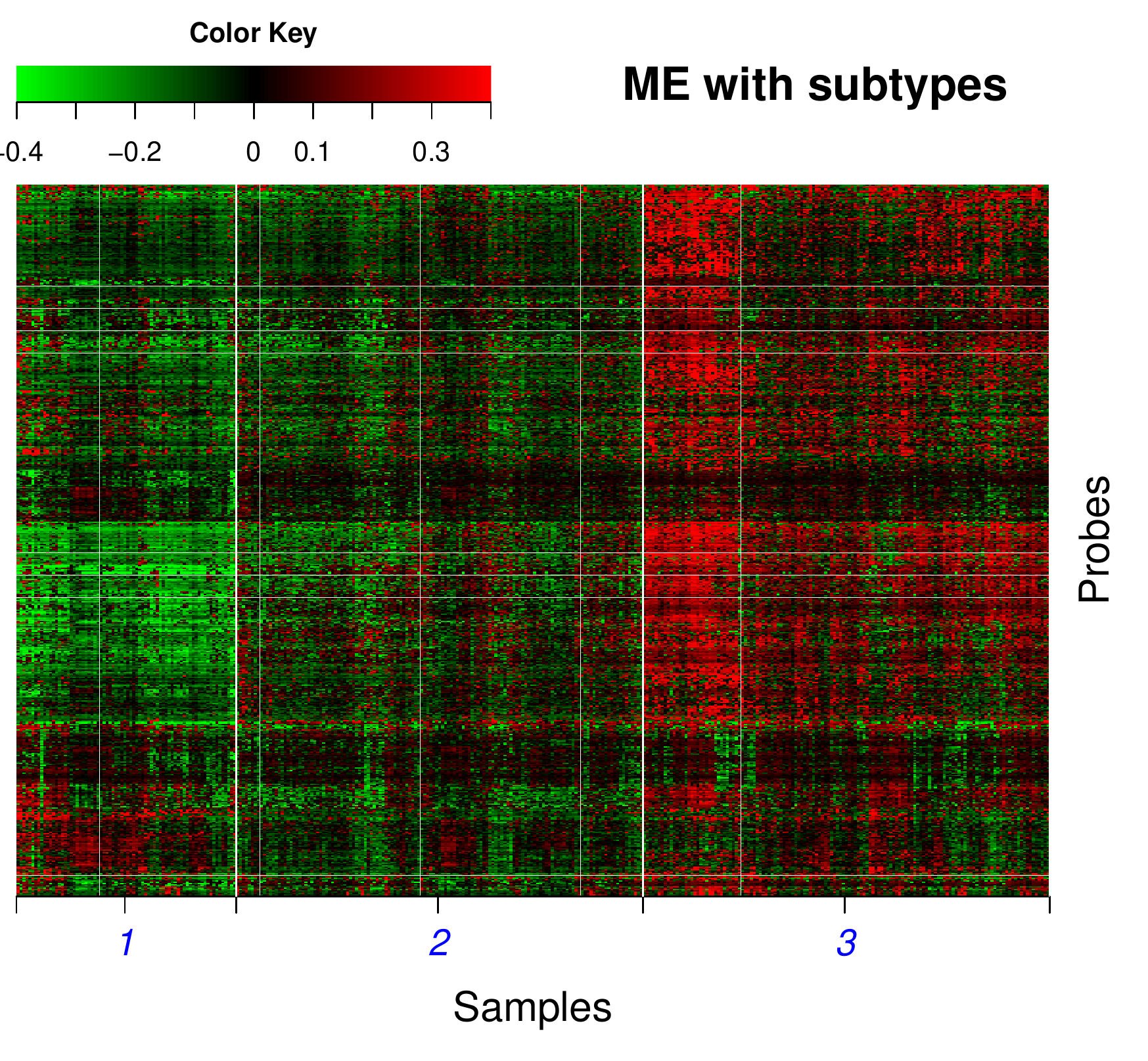}}
\caption{Heatmap of the ME data;  samples are grouped by their ME-specific cluster. }
\label{fig:MethHeatmap}
\end{center}
\vskip -0.2in
\end{figure}

\begin{figure}[!ht]
\vskip 0.2in
\begin{center}
\centerline{\includegraphics[width=\columnwidth, trim = 0mm 5mm 0mm 7mm, clip = TRUE]{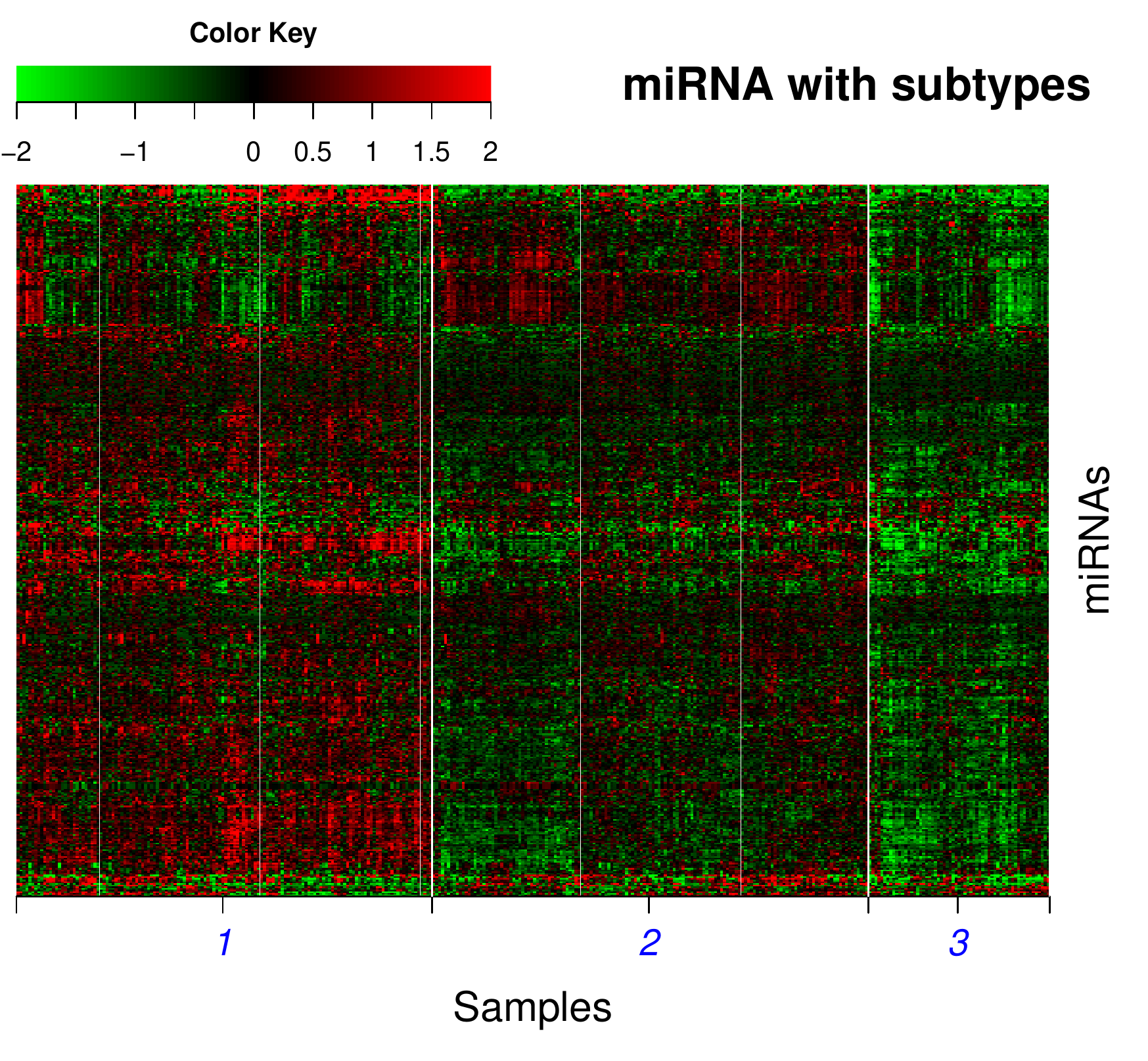}}
\caption{Heatmap of the miRNA data;  samples are grouped by their miRNA-specific cluster. }
\label{fig:miRNAHeatmap}
\end{center}
\vskip -0.2in
\end{figure}

\begin{figure}[!ht]
\vskip 0.2in
\begin{center}
\centerline{\includegraphics[width=\columnwidth, trim = 0mm 5mm 0mm 7mm, clip = TRUE]{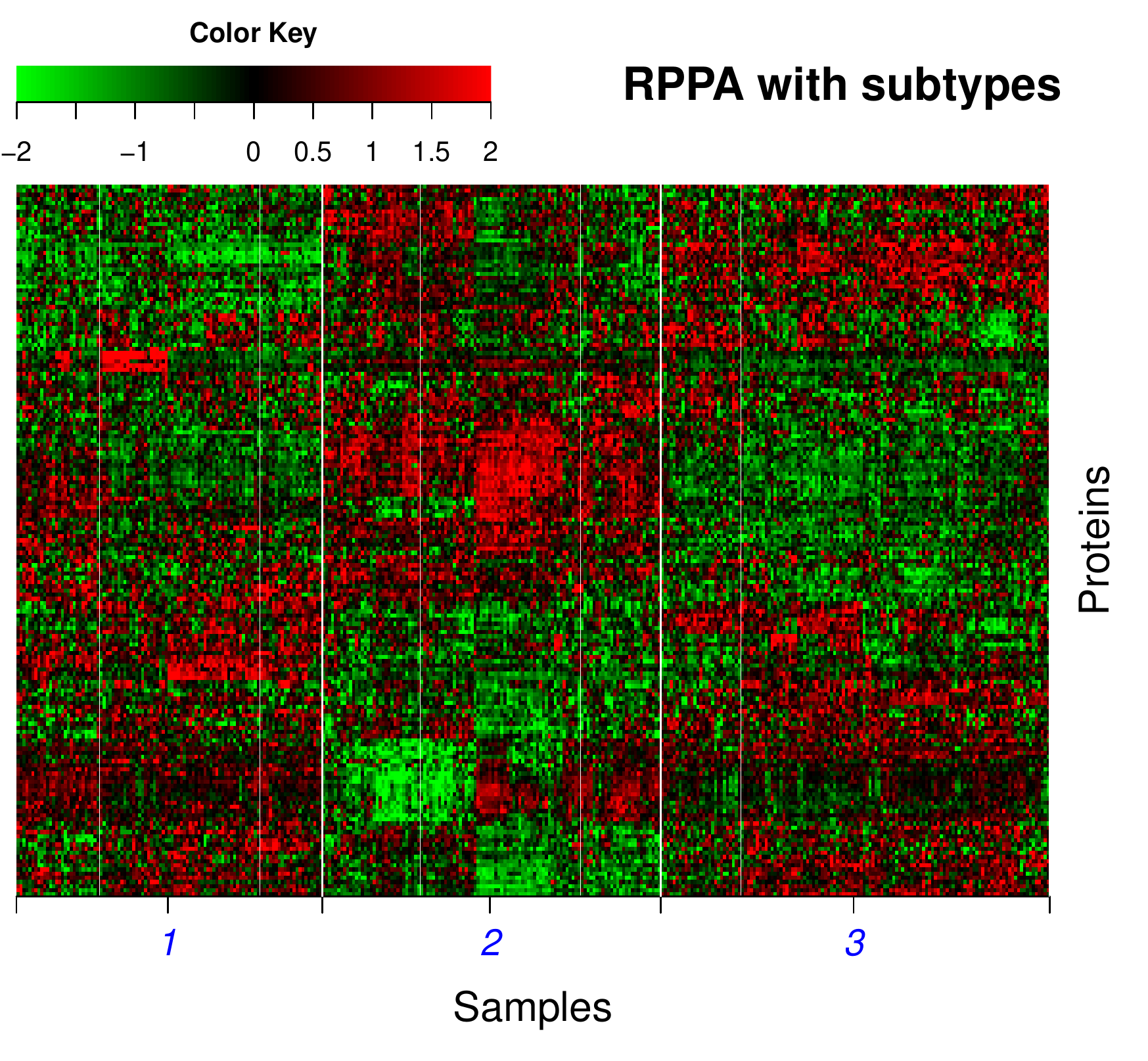}}
\caption{Heatmap of the RPPA data;  samples are grouped by their RPPA-specific cluster.}
\label{fig:RPPAHeatmap}
\end{center}
\vskip -0.2in
\end{figure}

Figures~\ref{fig:GEHeatmap},~\ref{fig:MethHeatmap},~\ref{fig:miRNAHeatmap}, and~\ref{fig:RPPAHeatmap} show heatmaps of the GE, ME, miRNA, and RPPA datasets, respectively.  The processed data that was used for clustering is shown in each case.  Samples (columns) are grouped by their relevant source-specific cluster.  For each heatmap a cluster effect is apparent in a large number of variables (rows).   

\bibliography{example_paper}
\bibliographystyle{icml2013}

\end{document}